\documentclass[runningheads]{llncs}

\usepackage{url}
\usepackage{bbold}
\usepackage{multirow}
\usepackage{graphicx}
\usepackage{booktabs}
\usepackage{amsfonts}
\usepackage{amsmath}
\DeclareMathOperator*{\argmax}{arg\,max}

\usepackage{enumitem}
\usepackage[sort]{cite}
\usepackage[hidelinks]{hyperref}
\usepackage[noend,algo2e,boxruled,linesnumbered]{algorithm2e}
\usepackage[misc]{ifsym}
\usepackage[dvipsnames]{xcolor}
\usepackage[export]{adjustbox}

\usepackage{scalerel}
\usepackage{tikz}
\usetikzlibrary{svg.path}

\definecolor{orcidlogocol}{HTML}{A6CE39}
\tikzset{
  orcidlogo/.pic={
    \fill[orcidlogocol] svg{M256,128c0,70.7-57.3,128-128,128C57.3,256,0,198.7,0,128C0,57.3,57.3,0,128,0C198.7,0,256,57.3,256,128z};
    \fill[white] svg{M86.3,186.2H70.9V79.1h15.4v48.4V186.2z}
                 svg{M108.9,79.1h41.6c39.6,0,57,28.3,57,53.6c0,27.5-21.5,53.6-56.8,53.6h-41.8V79.1z M124.3,172.4h24.5c34.9,0,42.9-26.5,42.9-39.7c0-21.5-13.7-39.7-43.7-39.7h-23.7V172.4z}
                 svg{M88.7,56.8c0,5.5-4.5,10.1-10.1,10.1c-5.6,0-10.1-4.6-10.1-10.1c0-5.6,4.5-10.1,10.1-10.1C84.2,46.7,88.7,51.3,88.7,56.8z};
  }
}

\newcommand\orcidicon[1]{\href{https://orcid.org/#1}{\mbox{\scalerel*{
    \begin{tikzpicture}[yscale=-1,transform shape]
    \pic{orcidlogo};
    \end{tikzpicture}
}{|}}}}

\newcommand{\lime}[1]{{\scshape lime}}
\newcommand{\anchor}[1]{{\scshape anchor}}
\newcommand{\maple}[1]{{\scshape maple}}
\newcommand{\shap}[1]{{\scshape shap}}
\newcommand{\lace}[1]{{\scshape lace}}
\newcommand{\brl}[1]{{\scshape brl}}
\newcommand{\rgc}[1]{{\scshape rgc}}
\newcommand{\soc}[1]{{\scshape soc}}
\newcommand{\lore}[1]{{\scshape lore}}

\newcommand{\bce}[1]{{\scshape bce}}
\newcommand{\bcet}[1]{{\scshape bce-t}}
\newcommand{\bcen}[1]{{\scshape bce-n}}
\newcommand{\bceb}[1]{{\scshape bce-b}}
\newcommand{\bces}[1]{{\scshape bce-s}}
\newcommand{\rce}[1]{{\scshape ece}}
\newcommand{\rces}[1]{{\scshape ece-s}}
\newcommand{\rceb}[1]{{\scshape ece-b}}
\newcommand{\rcet}[1]{{\scshape ece-t}}
\newcommand{\rcen}[1]{{\scshape ece-n}}
\newcommand{\rceh}[1]{{\scshape ece-h}}

\newcommand{\csg}[1]{{\scshape csg}}

\newcommand{\dice}[1]{{\scshape dice}}
\newcommand{\dce}[1]{{\scshape dce}}
\newcommand{\cem}[1]{{\scshape cem}}
\newcommand{\ceml}[1]{{\scshape ceml}}
\newcommand{\cegp}[1]{{\scshape cegp}}
\newcommand{\chvae}[1]{{\scshape chave}}
\newcommand{\face}[1]{{\scshape face}}
\newcommand{\piece}[1]{{\scshape piece}}
\newcommand{\vice}[1]{{\scshape vice}}
\newcommand{\mace}[1]{{\scshape mace}}
\newcommand{\macem}[1]{{\scshape macem}}
\newcommand{\brfo}[1]{{\scshape bf}}
\newcommand{\watc}[1]{{\scshape wach}}
\newcommand{\grace}[1]{{\scshape grace}}
\newcommand{\certifai}[1]{{\scshape certifai}}
\newcommand{\dace}[1]{{\scshape dace}}
\newcommand{\gic}[1]{{\scshape gic}}
\newcommand{\revise}[1]{{\scshape revise}}
\newcommand{\cbce}[1]{{\scshape cbce}}

\newcommand{\adult}[1]{\texttt{adult}}
\newcommand{\compas}[1]{\texttt{compas}}
\newcommand{\fico}[1]{\texttt{fico}}
\newcommand{\german}[1]{\texttt{german}}

\newcommand{\mnist}[1]{\texttt{mnist}}
\newcommand{\fashion}[1]{\texttt{fashion}}
\newcommand{\gunpoint}[1]{\texttt{gunpoint}}
\newcommand{\power}[1]{\texttt{power}}
\newcommand{\ecg}[1]{\texttt{ecg200}}

\makeatletter
\newcommand\footnoteref[1]{\protected@xdef\@thefnmark{\ref{#1}}\@footnotemark}
\makeatother

\begin{document}

\title{Ensemble of Counterfactual Explainers}
\titlerunning{Ensemble of Counterfactual Explainers}

\author{Riccardo Guidotti\orcidicon{0000-0002-2827-7613} \Letter \and 
    Salvatore Ruggieri \orcidicon{0000-0002-1917-6087}
}

\authorrunning{R. Guidotti and S. Ruggieri}

\tocauthor{Riccardo Guidotti, Salvatore Ruggieri}

\institute{University of Pisa, Italy, \email{\{name.surname\}@unipi.it}}

\maketitle 

\begin{abstract}
In eXplainable Artificial Intelligence (XAI), several counterfactual explainers have been proposed, each focusing on some desirable properties of counterfactual instances: minimality, actionability, stability, diversity, plausibility, discriminative power. We propose an ensemble of counterfactual explainers that boosts weak explainers, which provide only a subset of such properties, to a powerful method covering all of them. The ensemble runs weak explainers on a sample of instances and of features, and it combines their results by exploiting a diversity-driven selection function. The method is model-agnostic and, through a wrapping approach based on autoencoders, 
it is also data-agnostic.
\end{abstract}

\section{Introduction}
In eXplainable AI (XAI), several counterfactual explainers have been proposed, each focusing on some desirable properties of counterfactual instances.
Consider an instance $x$ for which a black box decision $b(x)$ has to be explained.
It should be possible to find various counterfactual instances $c$ (\textit{availability}) which are \textit{valid} (change the decision outcome, i.e., $b(c) \neq b(x)$), \textit{minimal} (the number of features changed in $c$ w.r.t.~$x$ should be as small as possible), \textit{actionable} (the feature values in $c$ that differ from $x$ should be controllable) and \textit{plausible} (the feature values in $c$ should be coherent with the reference population).
The counterfactuals found should be similar to $x$ (\textit{proximity}), but also different among each other (\textit{diversity}).
Also, they should exhibit a \textit{discriminative power} to characterize the black box decision boundary in the feature space close to $x$.
Counterfactual explanation methods should return similar counterfactuals for similar instances to explain (\textit{stability}). Finally, they must be fast enough (\textit{efficiency}) to allow for interactive usage.

In the literature, these desiderata for counterfactuals are typically modeled through an optimization problem~\cite{mothilal2019explaining}, which, on the negative side, favors only a subset of the properties above.
We propose here an \textit{ensemble of counterfactual explainers} (\rce{}) that, as in the case of ensemble of classifiers, boosts weak explainers to a powerful method covering all of the above desiderata. 
The ensemble runs \textit{base counterfactual explainers} (\bce{}) on a sample of instances and of features, and it combines their results by exploiting a diversity-driven selection function. 
The method is model-agnostic and, through a wrapping approach based on encoder/decoder functions, it is also data-agnostic. We will be able to reason uniformly on counterfactuals for tabular data, images, and time series. An extensive experimentation is presented to validate the approach. We compare with state-of-the-art explanation methods on several metrics from the literature.

\section{Related Work}
\label{sec:related}
Research on XAI has flourished over the last few years~\cite{guidotti2018survey}.
Explanation methods can be categorized as: \emph{(i)} \textit{intrinsic} vs \textit{post-hoc}, depending on whether the AI model is directly interpretable, or if the explanation is computed for a given black box model;  
\emph{(ii)} \textit{model-specific} vs \textit{model-agnostic}, depending on whether the approach requires access to the internals of the black box model; \emph{(iii)} \textit{local} or \textit{global}, depending on whether the explanation regards a specific instance, or the overall logic of the black box.
Furthermore, explanation methods can be categorized w.r.t.~the type of explanation they return (factual or counterfactual) and w.r.t.~the type of data they work with. % (tabular data, images, text, time series, etc.).
We restrict to local and post-hoc methods returning counterfactual explanations, which is the focus of our proposal.

A recent survey of counterfactual explainers is~\cite{verma2020counterfactual}.
Most of the systems are data-specific and generate synthetic (\textit{exogenous}) counterfactuals. Some approaches search \textit{endogenous} counterfactuals in a given dataset~\cite{keane2020good} of instances belonging to the reference population. 
Exogenous counterfactuals may instead break known relations between features, producing unrealistic instances.
Early approaches generated exogenous counterfactuals by solving an optimization problem~\cite{mothilal2019explaining}.
In our proposal, we do not rely on this family of methods as they are typically computationally expensive.
Another family of approaches are closer to instance-based classification, and rely on a distance function among instances~\cite{keane2020good,laugel2018comparison}.
E.g.,~\cite{laugel2018comparison} grows a sphere around the instance to explain, stopping at the decision boundary of the black box. 
They are simple but effective, and the idea will be at the core of our base explainers.
Some approaches deal with high dimensionality of data through autoencoders~\cite{dhurandhar2018explanations}, which map instances into a smaller latent feature space. 
Search for counterfactuals is performed in the latent space, and then instances are decoded back to the original space. We rely on this idea to achieve  a data-agnostic approach.

\section{Problem Setting}
\label{sec:problem}
A \textit{classifier} $b$ is a function mapping an instance $x$ from a reference population in a feature space to a nominal value $y$ also called class value or decision, i.e., $b(x) = y$. 
%We use $b(X) = Y$ as a shorthand for $\{b(x) | x \in X\} = Y$.
The classifier $b$ is a \emph{black box} when its internals are either unknown to the observer or they are known but uninterpretable by humans. 
Examples include neural networks, SVMs, ensemble classifiers~\cite{guidotti2018survey}.

A \textit{counterfactual} of $x$ is an instance $c$ for which the decision of the black box differs from the one of $x$, i.e., such that $b(c) \neq b(x)$. 
A counterfactual is \textit{actionable} if it belongs to the reference population. Since one may not have a complete specification of the reference population, a relaxed definition of actionability is to require the counterfactual to satisfy given constraints on its feature values. 
We restrict %in this paper 
to simple constraints $a_A(c, x)$ that hold iff $c$ and $x$ have the same values over for a set $A$ of \textit{actionable features}. Non-actionable features (such as age, gender, race) %are those that 
cannot be changed when searching for a counterfactual. 

A \textit{$k$-counterfactual explainer} is a function $f_k$ returning a set $C=\{c_1, \dots, c_h\}$ of $h \leq k$ actionable counterfactuals for a given instance of interest $x$, a black box $b$, a set $X$ of known instances from the reference population, and a set $A$ of actionable features, i.e., $f_k(x, b, X, A) = C$. For endogenous approaches, $C \subseteq X$. 
%For machine learning developers adopting explainability methods for  debugging black boxes, the natural choice of $X$ is the training set. 
A counterfactual explainer is model-agnostic (resp., data-agnostic) if the definition of $f_k$ does not depend on the internals of $b$ (resp., on the data type of $x$).
We consider the following data types: tabular data, time series and images.
For \textit{tabular data}, an instance $x = \{(a_1, v_1), \dots, (a_m, v_m) \}$ is a tuple of $m$ attribute-value pairs $(a_i, v_i)$, where $a_i$ is a feature (or attribute) and $v_i$ is a value from the domain of $a_i$. 
For example, $x = \{ (\mathit{age}, 22), (\mathit{sex}, \mathit{male}), (\mathit{income}, \mathit{800}) \}$. %, (\mathit{car},\mathit{no})\}$
The domain of a feature can be continuous (\textit{age}, \textit{income}), or categorical (\textit{sex}). %, \textit{car}).
For (univariate) \textit{time series}, an instance $x = \langle v_1, \dots, v_m \rangle$ is an ordered sequence of continuous values (e.g., the body temperature registered at hourly rate).
For \textit{images}, $x$ is a matrix in $\mathbb{R}^{m \times m}$ representing the intensity of the image pixels. 

\smallskip
\textit{Problem Statement.} We consider the problem of designing a $k$-counterfactual explainer satisfying a broad range of properties: availability, validity, actionability, plausibility, similarity, diversity, discriminative power, stability, efficiency.

\section{Ensemble of Explainers}
\label{sec:method}
Our proposal to the stated problem consists of  an ensemble of base explainers named \rce{} ({\scshape e}nsamble of {\scshape c}ounterfactual {\scshape e}xplainers).
Ensemble classifiers boost the performance of weak learner base classifiers by increasing the predictive power, or by reducing bias or variance. Similarly, we aim at improving base $k$-counterfactual explainers by combining them into an ensemble of explainers. 

\begin{algorithm2e}[t]
\scriptsize
    \caption{\rce{}}
    \label{alg:rce}
    \SetKwInOut{Input}{Input}
	\SetKwInOut{Output}{Output}
	\Input{$x$ - instance to explain,
    $b$ - black box,
    $X$ - known instances,\\
    $k$ - number of counterfactuals,
    $A$ - actionable features,
    $E$ - base explainers
    }
	\Output{$C$ - $k$-counterfactual set}
	\BlankLine
	$C \leftarrow \emptyset$;\tcp*[f]{\texttt{\scriptsize init. result set}}\\
	\For(\tcp*[f]{\texttt{\scriptsize for each base explainer}}){$f_k \in E$}{
	    $X' \leftarrow \mathcal{I}(X)$;\tcp*[f]{\texttt{\scriptsize sample instances}}\\
	    $A' \leftarrow \mathcal{F}(A)$;\tcp*[f]{\texttt{\scriptsize sample features}}\\
	    $C \leftarrow C \cup f_k(x, b, X', A')$;\tcp*[f]{\texttt{\scriptsize call base explainer}}\\
	}
    $C \leftarrow \mathcal{S}(x, C, k)$; \tcp*[f]{\texttt{\scriptsize select top $k$-counterfactuals}}\\
    \Return{$C$};\\ 
\end{algorithm2e}

The pseudo-code\footnote{Implementation and full set of parameters at \url{https://github.com/riccotti/ECE}} of \rce{} is shown in Alg.~\ref{alg:rce}.
It takes as input an instance to explain $x$, the black box to explain $b$, a set of known instances $X$, the number of required counterfactuals $k$, the set of actionable features $A$, a set of base $k$-counterfactual explainers $E$, and it returns (at most) $k$ counterfactuals $C$.
Base explainers are invoked on a sample without replacement $X'$ of instances from $X$ (line 3), and on a random subset $A'$ of the actionable features $A$ (line 4), as in Random Forests. All counterfactuals produced by the base explainers are collected in a set $C$ (line 5), from which $k$ counterfactuals are selected (line 6). 
Actionability of counterfactuals is guaranteed by the base explainers (or by filtering out non-actionable ones from their output).
Diversity is enforced by randomization (instance and feature sampling) as well as by tailored selection strategies. Stability is a result of combining multiple base explainers, analogously to the smaller variance of ensemble classification w.r.t.~the base classifiers. Moreover, if all base explainers are model-agnostic, this also holds for \rce{}.
%A few base explainers are introduced next, whilst selection strategies $\mathcal{S}$ are discussed in Sect.~\ref{sec:select}.
%Finally, Sect.~\ref{sec:ae} shows how is possible to apply \rce{} on any data type by wrapping it around encoding/decoding functions.

\subsection{Base Explainers}
\label{sec:bce}
%In the following, we write $X_{(\mathit{cond}(X, \ldots))}$ to denote the instance $x' \in X$ such that $\mathit{cond}(x', \ldots)$ holds. E.g., $X_{(b(X) \neq b(x))}$ is the set of instances $x' \in X$ such that $b(x') \neq b(x)$. We denote by $X_{[S]}$ the projection of the instances in $X$ over the features in $S$. E.g., $X_{[A]}$ projects over the actionable features $A$.
All \bce{}'s presented are parametric to a distance function $d()$ over the feature space. 
%We write $d_{[A]}(x_1, x_2)$ to denote the distance between $x_1$ and $x_2$ over the space of actionable features $A$. 
In the experiments, we adopt: for tabular data, a mixed distance weighting Euclidean distance for continuous features and the Jaccard dissimilarity for categorical ones; for images and times series, the Euclidean distance.

\smallskip
\textbf{Brute Force Explainer (}\bceb{}\textbf{).}
A brute force approach considers all subsets $\mathcal{A}$ of actionable features $A$ with cardinality at most $n$.
Also, for each actionable feature, an equal-width  binning into $r$ bins is computed, and for each bin the center value will be used as representative of the bin. The binning scheme considers only the known instances $X$ with black box decision different from $x$.
The brute force approach consists of generating all the possible variations of $x$ with respect to any of the subset in $\mathcal{A}$ by replacing an actionable feature value in $x$ with any representative value of a bin of the feature.
Variations are ranked according to their distance from $x$. 
For each such variation $c$, a $\mathit{refine}$ procedure implements a bisecting strategy of the features in $c$ which are different from $x$ while maintaining $b(c) \neq b(x)$. The procedure returns either a singleton with a counterfactual or an empty set (in case $b(c) = b(x)$). The aim of $\mathit{refine}$ is to improve similarity of the counterfactual with $x$. 
The procedure stops when $k$ counterfactuals have been found or there is no further candidate.
The greater are $n$ and $r$, the larger number of counterfactuals to choose from, but also 
the higher the computational complexity of the approach, which is $O({|A| \choose n} \cdot n \cdot r)$. 
%We will show in the experimental section that $n=2$ and $r \in \{5, 10, 20\}$ are sufficient for achieving good counterfactuals. 
\bceb{} tackles  minimization of changes and similarity, but not diversity.

%\smallskip
%\textbf{Nearest Neighbor Explainer (}\bcen{}\textbf{).}
%A nearest neighbor approach is designed as a simplified version of \cbce{}~\cite{keane2020good}.
%\bcen{} calculates the distances between $x$ and the instances in $X$ with black box decision different from $x$. %the one for $x$. 
%The distances are calculated over the space of actionable features.
%Candidate counterfactuals $c$ are sorted w.r.t.~the similarity with $x$. 
%Actionability is ensured by overwriting non-actionable features of $c$ with their value in $x$.
%If $c$ is a valid counterfactual, it is added to the result set.
%The procedure stops when $k$ counterfactuals have been found or there is no further candidate.
%A weakness of \bcen{} is the computational cost of computing distances, which is $O(|X| \cdot |A|)$. 
%Notice, however, that since \rce{}  performs a sampling of the actionable features, we have that $|A|$ ranges over reasonable small values.
%This approach only accounts for similarity and validity, but not for diversity and minimization of changes.

\smallskip
\textbf{Tree-based Explainer (}\bcet{}\textbf{).}
This proposal starts from a (surrogate/sha\-dow~\cite{guidotti2019factual}) decision tree $\mathcal{T}$ trained on $X$ to mime the black box behavior. 
Leaves in $T$ leading to predictions different from $b(x)$ can be exploited for building counterfactuals. 
Basically, the splits on the path from the root to one such leaf represent conditions satisfied by counterfactuals. 
To ensure actionability, only splits involving actionable constraints are considered. 
To tackle minimality, the filtered paths are sorted w.r.t.~the number of conditions not already satisfied by $x$. For each such path, we choose one instance $c$ from $X$ reaching the leaf and minimizing distance to $x$. Even though the path has been checked for actionable splits, the instance $c$ may still include changes w.r.t.~$x$ that are not actionable. For this, we overwrite non-actionable features. Since not all instances at a leaf have the same class as the one predicted at the leaf, we also have to check for validity before including $c$ in the result set.
The search over different paths of the decision tree allows for some diversity in the results, even though this cannot be explicitly controlled for.
The computational complexity requires both a decision tree construction and a number of distance calculations.

\smallskip
\textbf{Generative Sphere-based Explainer (}\bces{}\textbf{).}
The last base counterfactual explainer relies on a generative approach growing a \textit{sphere} of synthetic instances around $x$~\cite{laugel2018comparison}.
Instance are generated  in all directions of the feature space until the decision boundary of the black box $b$ is crossed and the closest counterfactual to $x$ is retrieved.
The sphere radius is initialized to a large value, and then it is decreased until the boundary is crossed. Next, a lower bound radius and an upper bound radius are determined such that the boundary of $b$ crosses the area of the sphere between the lower bound and the upper bound radii. 
In its original version, the growing spheres algorithm generates instances following a uniform distribution.
\bces{} adopts instead a \textit{Gaussian-Matched} generation~\cite{agustsson2017optimal}.
To ensure actionability, non-actionable features of generated instances are set as in $x$. 
Finally, \bces{} selects from the instances in the final ring the ones which are closest to $x$ and are valid. 
%The sphere-based approach accounts for diversity of counterfactuals by generating instances in a ring crossing the decision boundary of $b$.
The complexity of the approach depends on the distance of the decision boundary from $x$, which in turn determines the number of iterations needed to compute the final ring.

\subsection{Counterfactual Selection}
\label{sec:select}
The selection function $\mathcal{S}$ at line 5 of Alg.~\ref{alg:rce} selects $k$-counterfactuals from those returned by the base explainers. 
This problem can be formulated as maximizing an objective function over $k$-subsets of valid counterfactuals $C$.
We adopt a \textit{density-based} objective function: 
{\small \begin{equation*}
    \label{eq:dense_h} 
   \underset{S \subseteq C \wedge |S| \leq k}{\argmax} \  |\bigcup_{c \in S} \mathit{knn}_C(c)| - \lambda \sum_{c \in S} d(c, x) 
\end{equation*}}
It aims at  maximizing the difference between the size of neighborhood instances of the counterfactuals (a measure of diversity) and the total distance from $x$ (a measure of similarity) regularized by a parameter $\lambda$.
$\mathit{knn}_C(c)$ returns the $h$ most similar counterfactuals to $c$ among those in $C$. 
We adopt the Cost Scaled Greedy (\csg{}) algorithm \cite{DBLP:journals/corr/abs-2002-07782} for the above maximization problem. %Hyper-parameters $h$ an $\lambda$ are 

\subsection{Counterfactuals for Other Data Types}
\label{sec:ae}
We enable \rce{} to work on data types other than tabular data by wrapping it around two functions.
An \textit{encoder} $\zeta: \mathbb{D} {\rightarrow} \mathbb{R}^q$ that maps an instance from its actual domain $\mathbb{D}$ to a latent space of continuous features, and a \textit{decoder} $\eta: \mathbb{R}^q {\rightarrow} \mathbb{D}$ that maps an instance of the latent space back to the actual domain. Using such functions, any explainer $f_k(x, b, X, A)$ can be extended to the  domain $\mathbb{D}$ by invoking $\eta(f_k(\zeta(x), b', \zeta(X), A'))$ where the black box in the latent space is $b'(x) = b(\eta(x))$. The definition of the actionable features in the latent space $A'$ depends on the actual encoder and decoder.

Let us consider the image data type (for time series, the reasoning is analogous). 
A natural instantiation of the wrapping that achieves dimensionality reduction with a controlled loss of information consists in the usage of \textit{autoencoders} (AE) ~\cite{hinton2006reducing}.
An AE is a neural network composed by an encoder and a decoder which are trained simultaneously for learning a representation that reduces the dimensionality while minimizing the reconstruction loss.
%Similarly to~\cite{guidotti2020lasts}, we adopt AEs to lift a method designed for the latent space to work on the actual domain. 
A drawback of this approach is that we cannot easily map actionable feature in the actual domain to features in the latent space (this is a challenging research topic on its own). For this, we set $A'$ to be the whole set of latent features and hence, we are not able to deal with actionability constraints.

\section{Experiments}
\label{sec:experiments}

\begin{table}[t]
    \caption{Datasets description and black box accuracy. $n$ is the no. of instances. $m$ is the no. of features. $m_{con}$ and $m_{cat}$ are the no. of continuous and categorical features respectively. $m_{act}$  is the no. of actionable features. $m_{1h}$ is the total no. of features after one-hot encoding. Rightmost columns report classification accuracy: NN stands for DNN for tabular data, and for CNN for images and time series.}
    \label{tab:datasets}
\centering
    \begin{tabular}{|cc|ccccccc|cc|} %c|}
    \hline
    \multicolumn{2}{|c|}{Dataset} & $n$ & $m$ & $m_{con}$ & $m_{cat}$ & $m_{act}$ & $m_{1h}$ & $l$ & RF & NN \\ %& LGBM \\
     \hline
    \multirow{4}{*}{\rotatebox[origin=c]{90}{\texttt{tabular}}} & \adult{} & 32,561 & 12 & 4 & 8 & 5 & 103 & 2 & .85 & .84 \\ %& .??\\
    & \compas{} & 7,214 & 10 & 7 & 3 & 7 & 17 & 3 & .56 & .61 \\ % & .??\\
    & \fico{} & 10,459 & 23 & 23 & 0 & 22 & - & 2 & .68 & .67 \\ %& .??\\
    & \german{} & 1,000 & 20 & 7 & 13 & 13 & 61 & 2 & .76 & .81 \\ %& .??\\
    \hline
    \multirow{2}{*}{\rotatebox[origin=c]{90}{\texttt{img}}} & \mnist{} & 60k & $28\times28$ & all & 0 & all & - & 10 & - & .99 \\ %& .??\\
    & \fashion{} & 60k & $28\times28$ & all & 0 & all & - & 10 & - & .97 \\ %& .??\\
    \hline
    \multirow{3}{*}{\rotatebox[origin=c]{90}{\texttt{ts}}} & \gunpoint{} & 250 & 150 & all & 0 & all & - & 2 & - & .72 \\ %& .??\\
    & \power{} & 1,096 & 24 & all & 0 & all & - & 2 & - & .98 \\ %& .??\\
    & \ecg{} & 200 & 96 & all & 0 & all & - & 2 & - & .76 \\ 
    \hline
    \end{tabular}
\end{table}

\textbf{Experimental Settings.} We consider a few datasets widely adopted as benchmarks in the literature (see Table~\ref{tab:datasets}). 
There are three time series datasets, two image datasets, and four tabular datasets. 
For each tabular dataset, we have selected the set $A$ of actionable features, as follows.
 \adult{}: age, education, marital status, relationship, race, sex, native country;
\compas{}: age, sex, race;
\fico{}: external risk estimate;
\german{}: age, people under maintenance, credit history, purpose, sex, housing, foreign worker.

For each dataset, we trained and explained the following black box classifiers: Random Forest (RF) as implemented by \textit{scikit-learn}, and Deep Neural Networks (DNN) implemented by \textit{keras} for tabular datasets,  and  Convolutional Neural Networks (CNNs) implemented with \textit{keras} for images and time series.
We split tabular datasets into a 70\% partition used for the training and 30\% used for the test, while image and time series datasets are already released in partitioned files.
For each black-box and for each dataset, we performed on the training set a random search with a 5-fold cross-validation for finding the best parameter setting.
The classification accuracy on the test set is shown in Table~\ref{tab:datasets} (right).

We compare our proposal against competitors from the state-of-the-art offering a software library
that is updated and easy to use.
\dice{}~\cite{mothilal2019explaining} handles categorical features, actionability, and allows for specifying the number $k$ of counterfactuals to return. 
However, it is not model-agnostic as it only deals with differentiable models such as DNNs.
The \textit{FAT}~\cite{sokol2019fat} library implements a brute force (\brfo{}) counterfactual approach.
It handles categorical data but not the number $k$ of desired counterfactuals nor actionability.
The \textit{ALIBI} library implements the counterfactual explainers \cem{}~\cite{dhurandhar2018explanations,luss2019generating}, \cegp{}~\cite{van2019interpretable} and \watc{}~\cite{wachter2017counterfactual}.
All of them are designed to explain DNNs, do not handle categorical features and return a single counterfactual, but it is possible to enforce actionability by specifying the admissible feature ranges. 
Finally, \ceml{}~\cite{ceml} is a model-agnostic toolbox for computing counterfactuals based on optimization that does not handle categorical features and returns a single counterfactual.
We also re-implemented the case-based counterfactual explainer (\cbce{}) from~\cite{keane2020good}.
For each tool, we use the default settings offered by the library or suggested in the reference paper.
For each dataset, we explain 100 instances $x$ from the test set. The set $X$ of known instances in input to the explainers is the training set of the black box.
We report aggregated results as means over the 100 instances, datasets and black boxes.

\medskip {\bf Evaluation Metrics.}
We evaluate the performances of counterfactual explainers under various perspectives~\cite{mothilal2019explaining}.
The measures reported in the following are stated for a single instance $x$ to be explained, and considering the returned $k$-counterfactual set $C = f_k(x, b, X, A)$.
The metrics are obtained as the mean value of the measures over all $x$'s to explain.

\textit{Size.} The number of counterfactuals $|C|$ can be lower than $k$. 
We define $\mathit{size} = |C|/k$.
The higher the better.
Recall that by definition of a $k$-counterfactual explainer, any $c \in C$ is valid, i.e.,~$b(c) \neq b(x)$.

\textit{Actionability.} It accounts for the counterfactuals in $C$ that can be realized: $\mathit{act} = |\{c \in C\ | \; a_A(c, x)\}|/k$. The higher the better. 

\textit{Implausibility.} It accounts for how close are counterfactuals to the reference population. It is the average distance of $c \in C$ from the closest instance in the known set $X$.
The lower the better.
{\small $$\mathit{impl} = \frac{1}{|C|}\sum_{c \in C} \min_{x \in X} d(c, x)$$}
\textit{Dissimilarity.} It measures the proximity between $x$ and the counterfactuals in $C$. The lower the better.
We measure it in two fashions. 
The first one, named $\mathit{dis_{dist}}$, is the average distance between $x$ and the counterfactuals in $C$. % where the usage of different distance functions $d$ can return different results.
The second one, $\mathit{dis_{count}}$, quantifies the average number of features changed between a counterfactual $c$ and $x$.
%The $\mathbb{1}_{\mathit{cond}}$ operator returns $1$ if $\mathit{cond}$ is true, and $0$ otherwise.
Let $m$ be the number of features.
{\small $$\mathit{dis_{dist}} = \frac{1}{|C|}\sum_{c \in C}d(x, c) \quad \quad
\mathit{dis_{count}} = \frac{1}{|C|m}\sum_{c \in C}\sum_{i=1}^m\mathbb{1}_{c_i \neq x_i}$$}
\textit{Diversity.} It accounts for a diverse set of counterfactuals, where different actions can be taken to recourse the decision of the black box. The higher the better.  
We denote by $\mathit{div_{dist}}$ the average distance between the counterfactuals in $C$, and by $\mathit{div_{count}}$ the average number of different features between the counterfactuals.
{\small $$\mathit{div_{dist}} = \frac{1}{|C|^2}\sum_{c \in C}\sum_{c' \in C}d(c, c') \quad \quad
\mathit{div_{count}} = \frac{1}{|C|^2m}\sum_{c \in C}\sum_{c' \in C}\sum_{i=1}^m\mathbb{1}_{c_i \neq c'_i}$$}

\textit{Discriminative Power.} It measures the ability to distinguish through a naive approach between two different classes only using the counterfactuals in $C$. 
In line with~\cite{mothilal2019explaining}, we implement it as follows.
The sets $X_{=} \subset X$ and $X_{\neq} \subset X$ such that $b(X_{=}) = b(x)$ and $b(X_{\neq}) \neq b(x)$ are selected such that the instances in $X_{=}, X_{\neq}$ are the $k$ closest to $x$. 
Then we train a simple 1-Nearest Neighbor (1NN) classifier using $C \cup \{x\}$ as training set, and $d$ as distance function.
The choice of 1NN is due to its simplicity and connection to human decision making starting from examples.
We classify the instances in $X_{=} \cup X_{\neq}$ and we use the accuracy of the 1NN as \textit{discriminative power} ($\mathit{dipo}$).

\textit{Instability.} It measures to which extent the counterfactuals $C$ are close to the ones obtained for the closest instance to $x$ in $X$ with the same black box decision. The rationale is that similar instances should obtain similar explanations~\cite{guidotti2019stability}. 
The lower the better.
{\small $$\mathit{inst} = \frac{1}{1+d(x, x')}\frac{1}{|C||C'|}\sum_{c \in C}\sum_{c' \in C'} d(c, c')$$
with $x'=\mathit{argmin}_{x_1 \in X\setminus\{x\}, b(x_1) = b(x)}\,d(x, x_1)$ and $C' = f_k(x', b, X, A)$.}

\textit{Runtime.} It measures the elapsed time required by the explainer to compute the counterfactuals. The lower the better. Experiments were performed on Ubuntu 20.04 LTS, 252 GB RAM, 3.30GHz x 36 Intel Core i9.

\smallskip
In line with~\cite{wachter2017counterfactual,mothilal2019explaining}, in the above evaluation measures, we adopt as distance $d$ the following mixed distance:
{\small $$d(a, b) = \frac{1}{m_{con}}\sum_{i \in \mathit{con}}\frac{|a_i - b_i|}{\mathit{MAD}_i} + \frac{1}{m_{cat}} \sum_{i \in \mathit{cat}}\mathbb{1}_{a_i \neq b_i}$$}
where $\mathit{con}$ (resp., $\mathit{cat}$) is the set of continuous (resp., categorical) feature positions.
Such a distance is not necessarily the one used by the compared explainers. In particular, it substantially differs from the one used by \rce{}.

\medskip {\bf Parameter Tuning.}
From an experimental analysis (not reported here) of the  impact of the components of \rce{}, we set: for \bceb{}, $r=10$ and $n=1$; and for \rce{}, $|E|=10$ base explainers chosen uniformly random.

\begin{figure*}[t]
    %\centering
    \noindent
    \hspace{-0.55cm}
    \includegraphics[trim = 0mm 0mm 0mm 0mm, clip,width=0.25\linewidth]{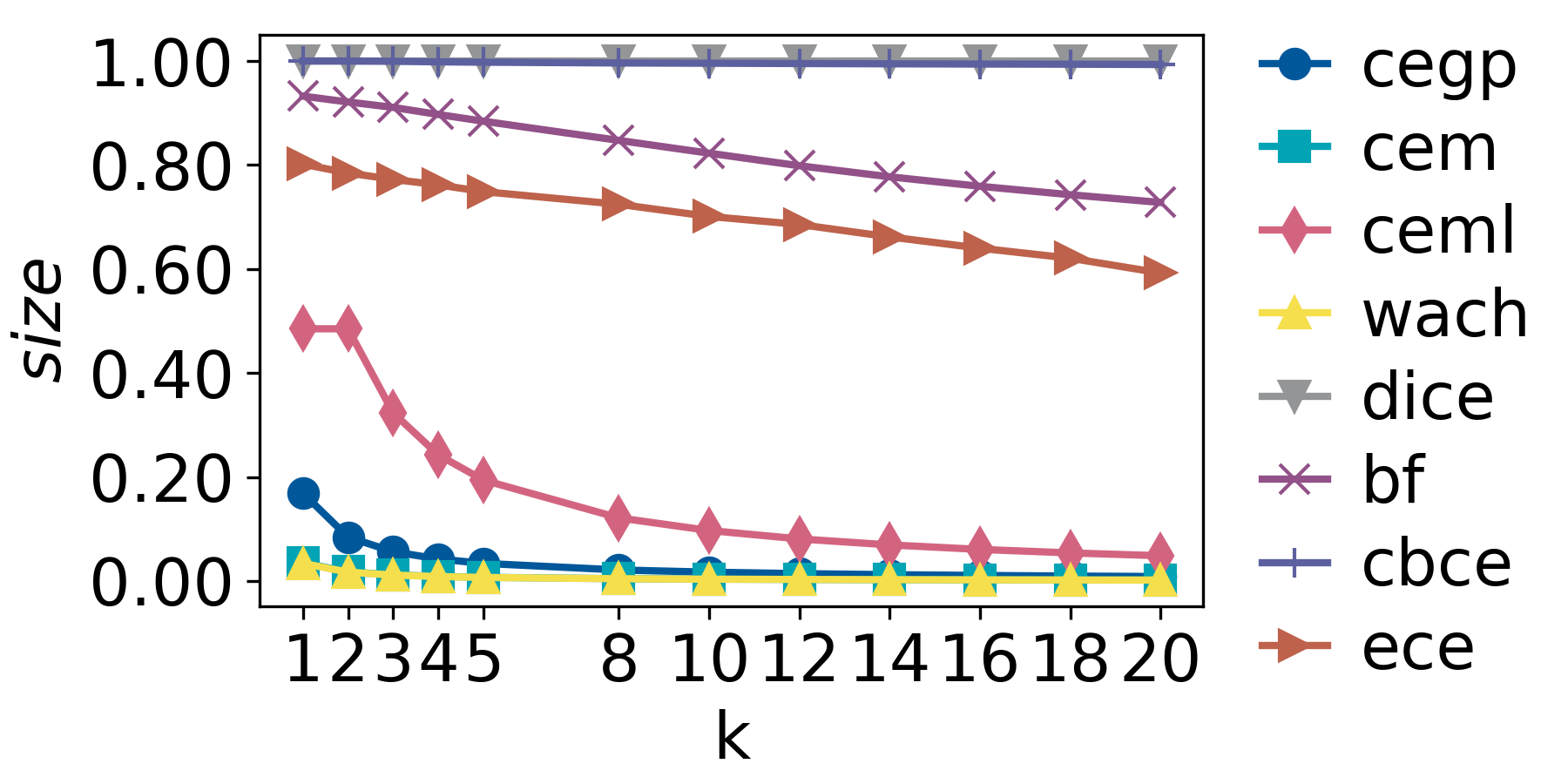}%
    \hspace{-0.65cm}\includegraphics[trim = 0mm 0mm 0mm 0mm, clip,width=0.25\linewidth]{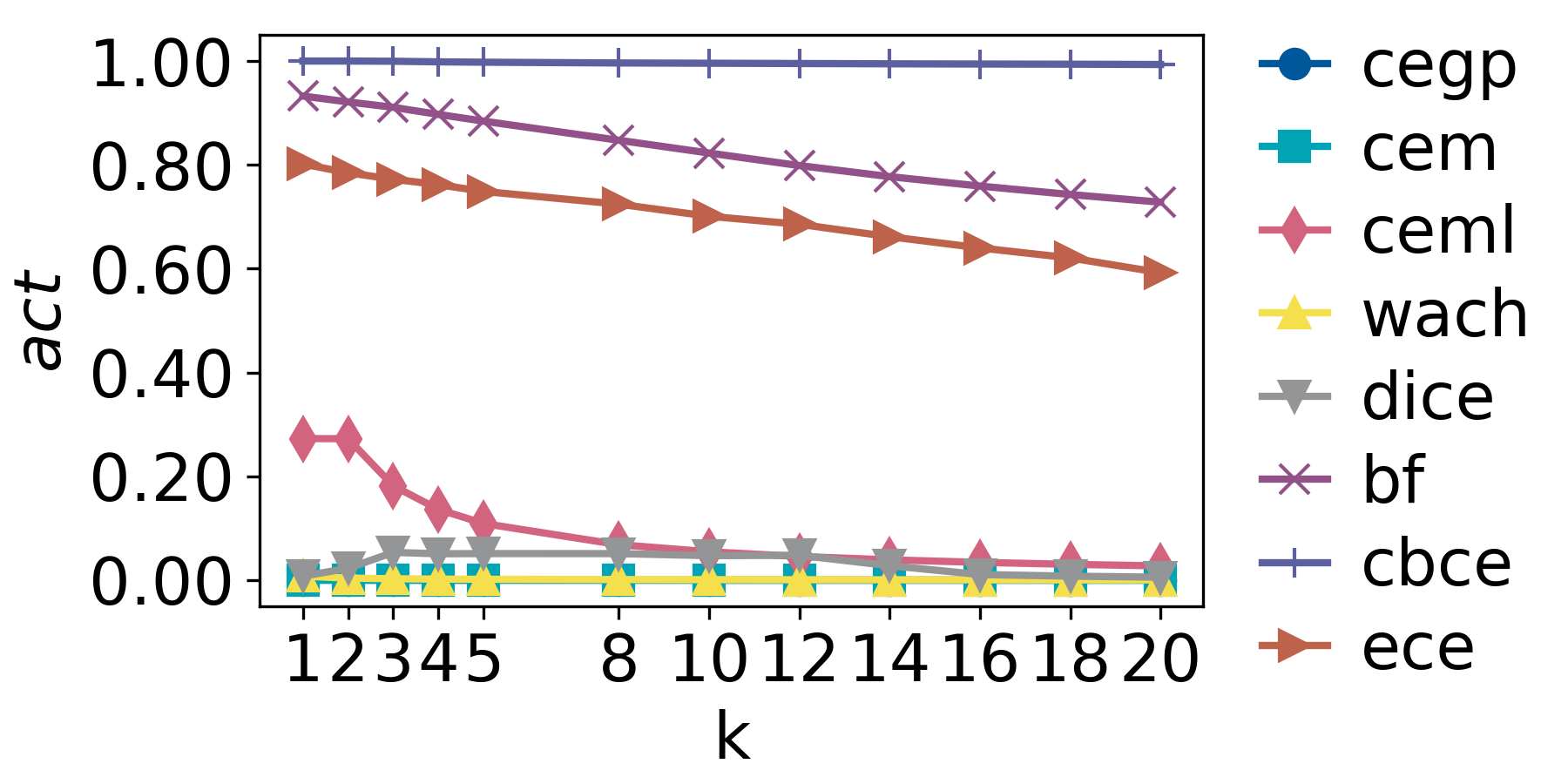}%
    \hspace{-0.65cm}\includegraphics[trim = 0mm 0mm 0mm 0mm, clip,width=0.25\linewidth]{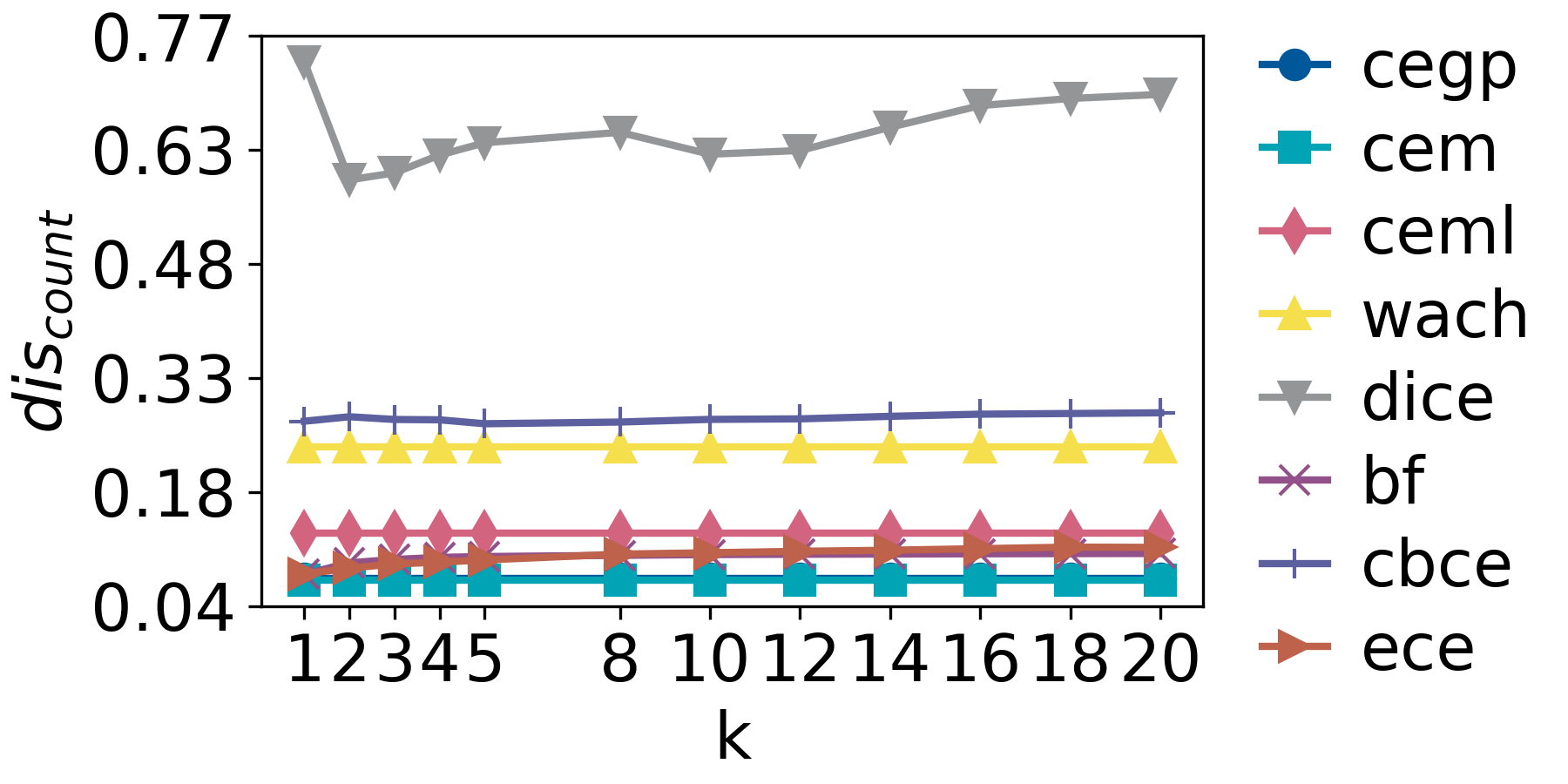}%
    \hspace{-0.65cm}\includegraphics[trim = 0mm 0mm 0mm 0mm, clip,width=0.25\linewidth]{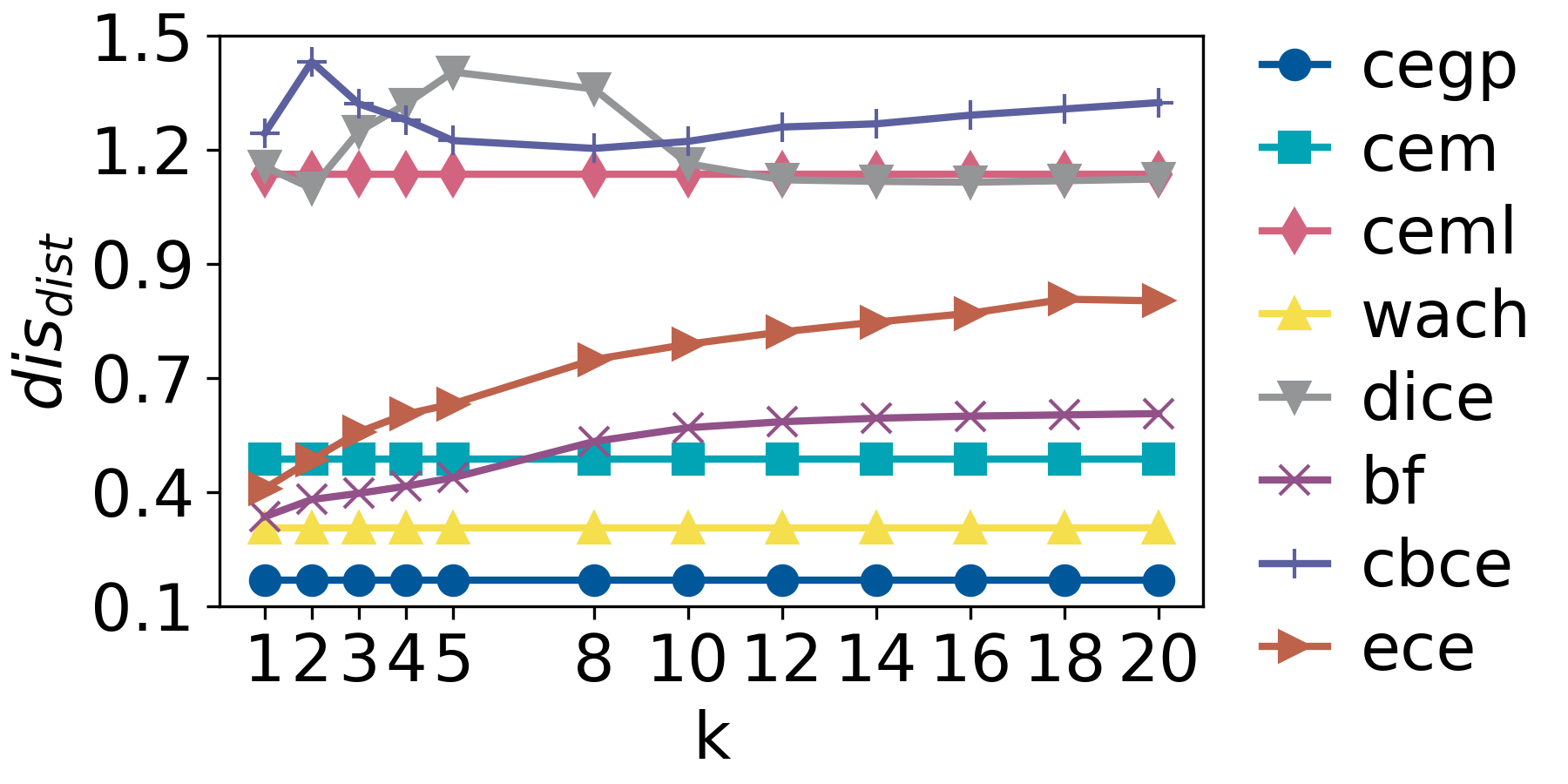}%
    \hspace{-0.65cm}\includegraphics[trim = 0mm 0mm 0mm 0mm, clip,width=0.25\linewidth]{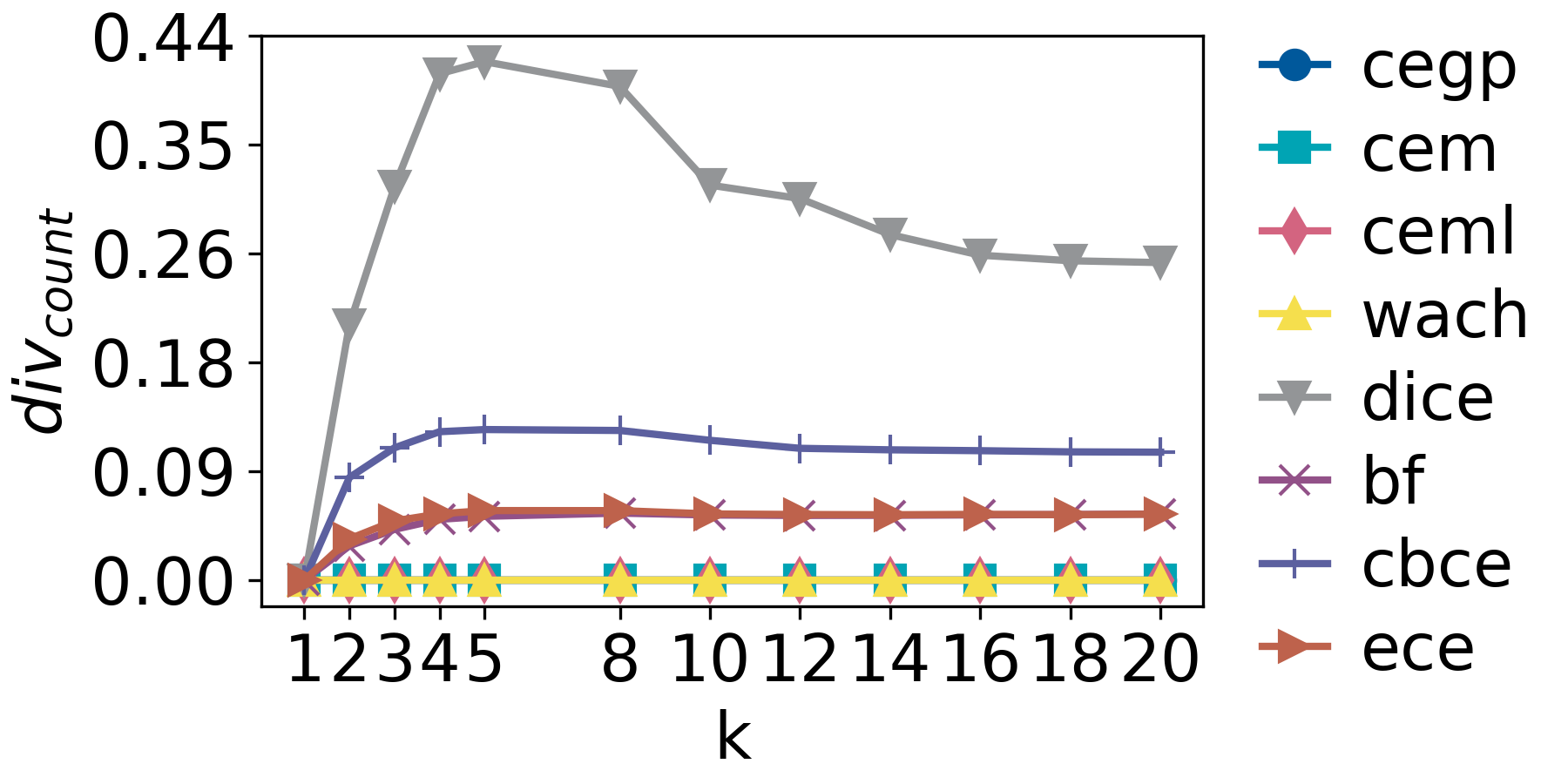} 
    
    %\noindent
    \hspace{-0.55cm}
    \includegraphics[trim = 0mm 0mm 0mm 0mm, clip,width=0.25\linewidth]{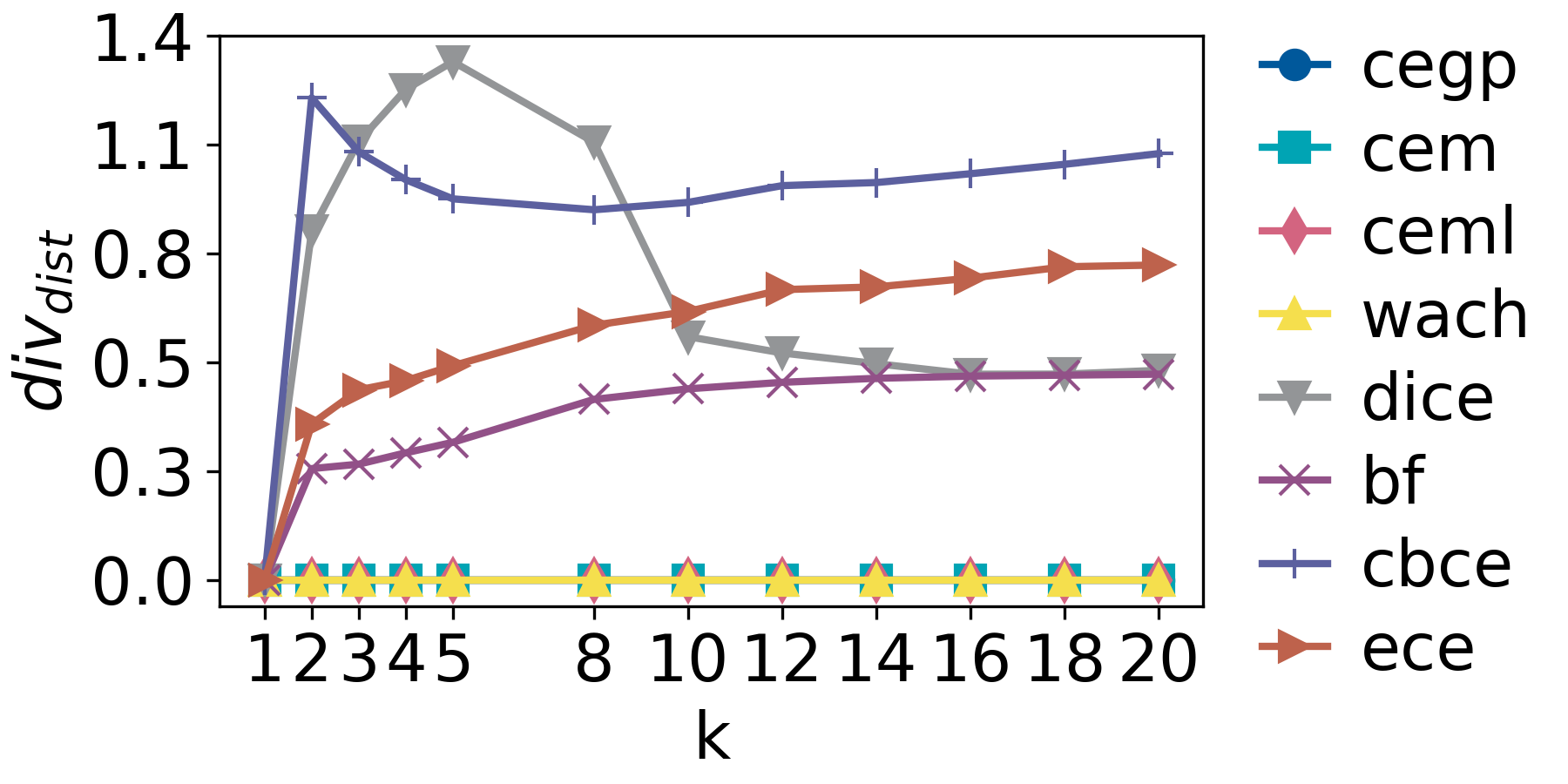}%
    \hspace{-0.65cm}\includegraphics[trim = 0mm 0mm 0mm 0mm, clip,width=0.25\linewidth]{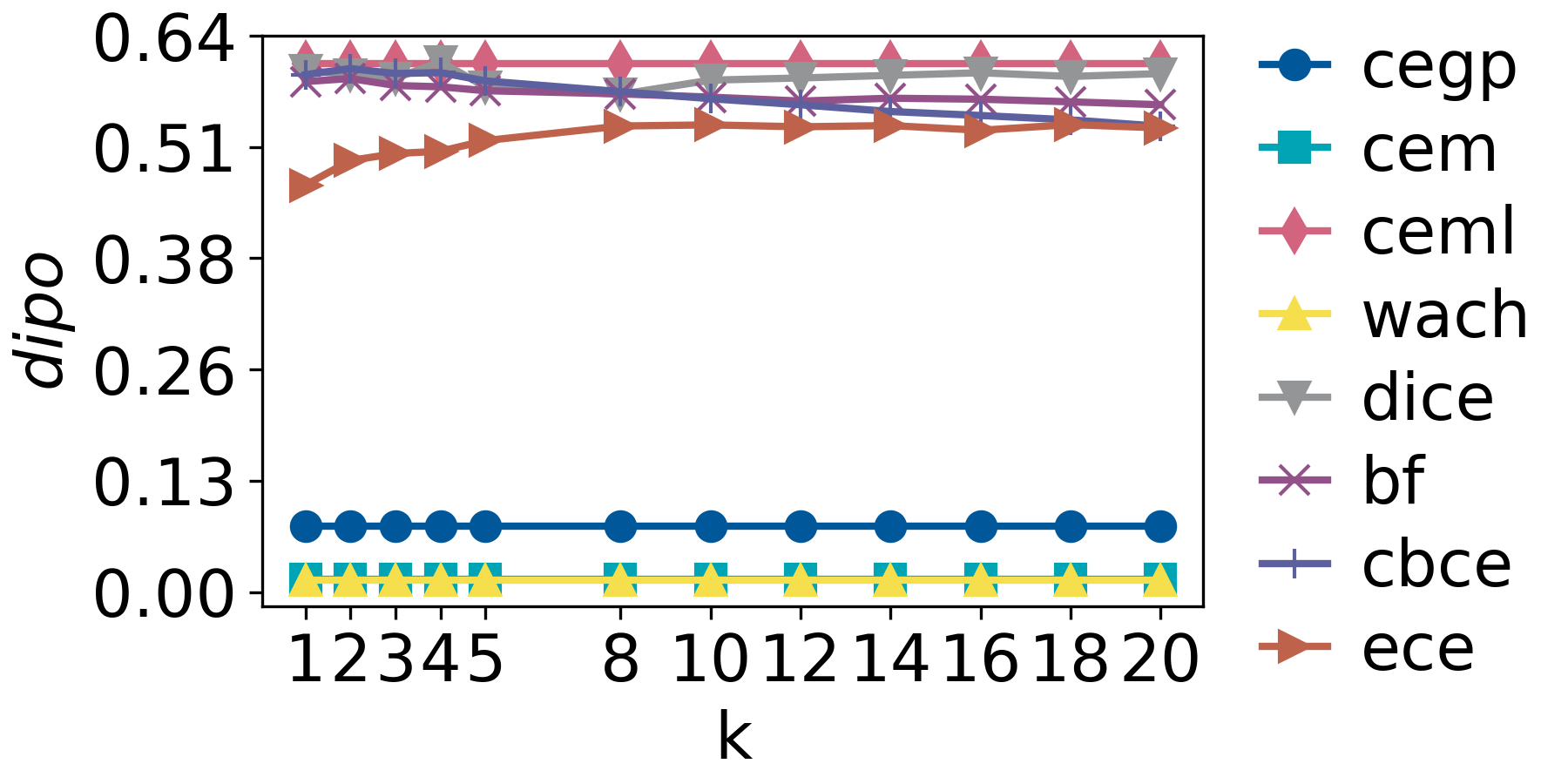}%
    \hspace{-0.65cm}\includegraphics[trim = 0mm 0mm 0mm 0mm, clip,width=0.25\linewidth]{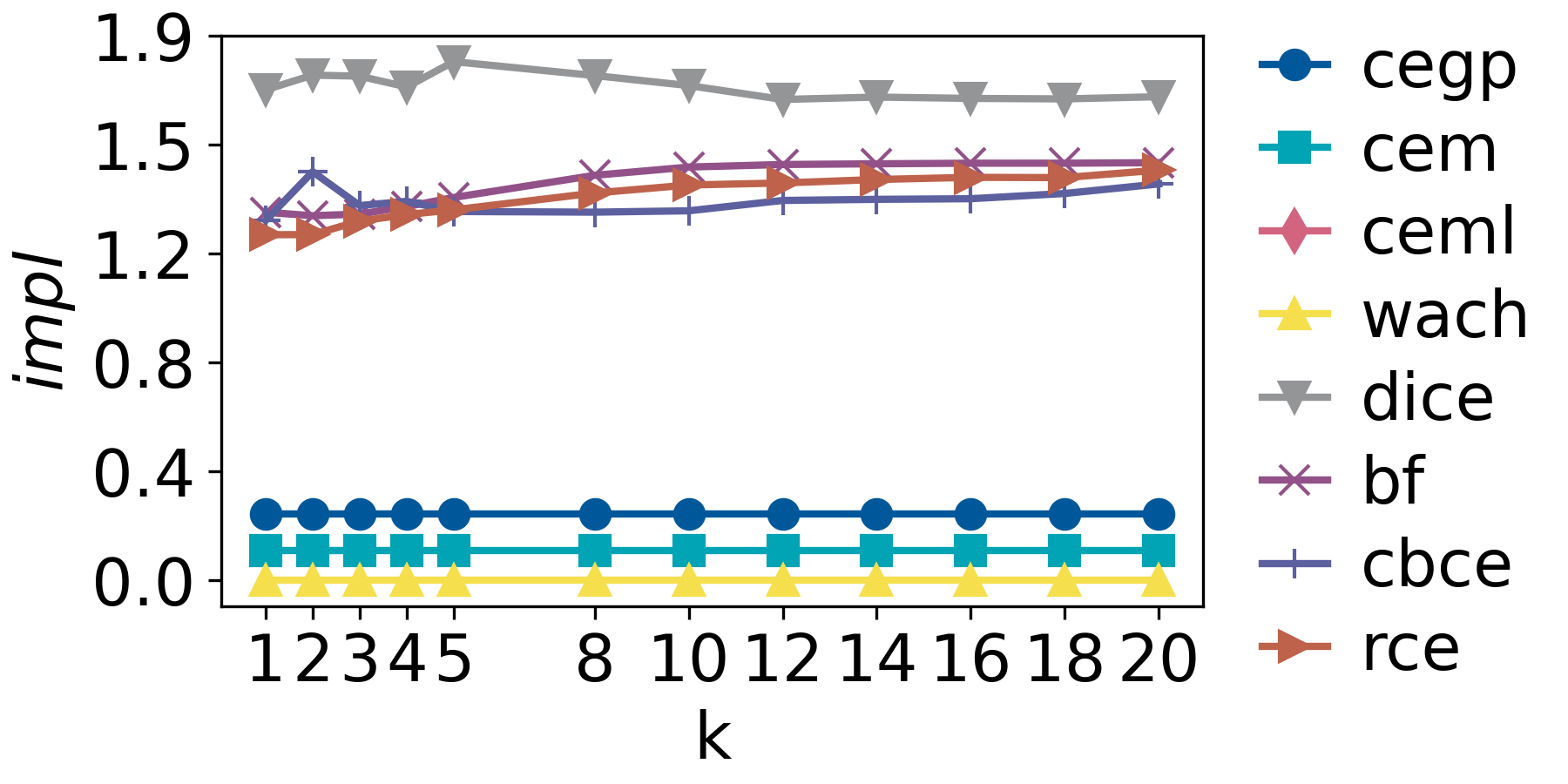}
    \hspace{-0.65cm}\includegraphics[trim = 0mm 0mm 0mm 0mm, clip,width=0.25\linewidth]{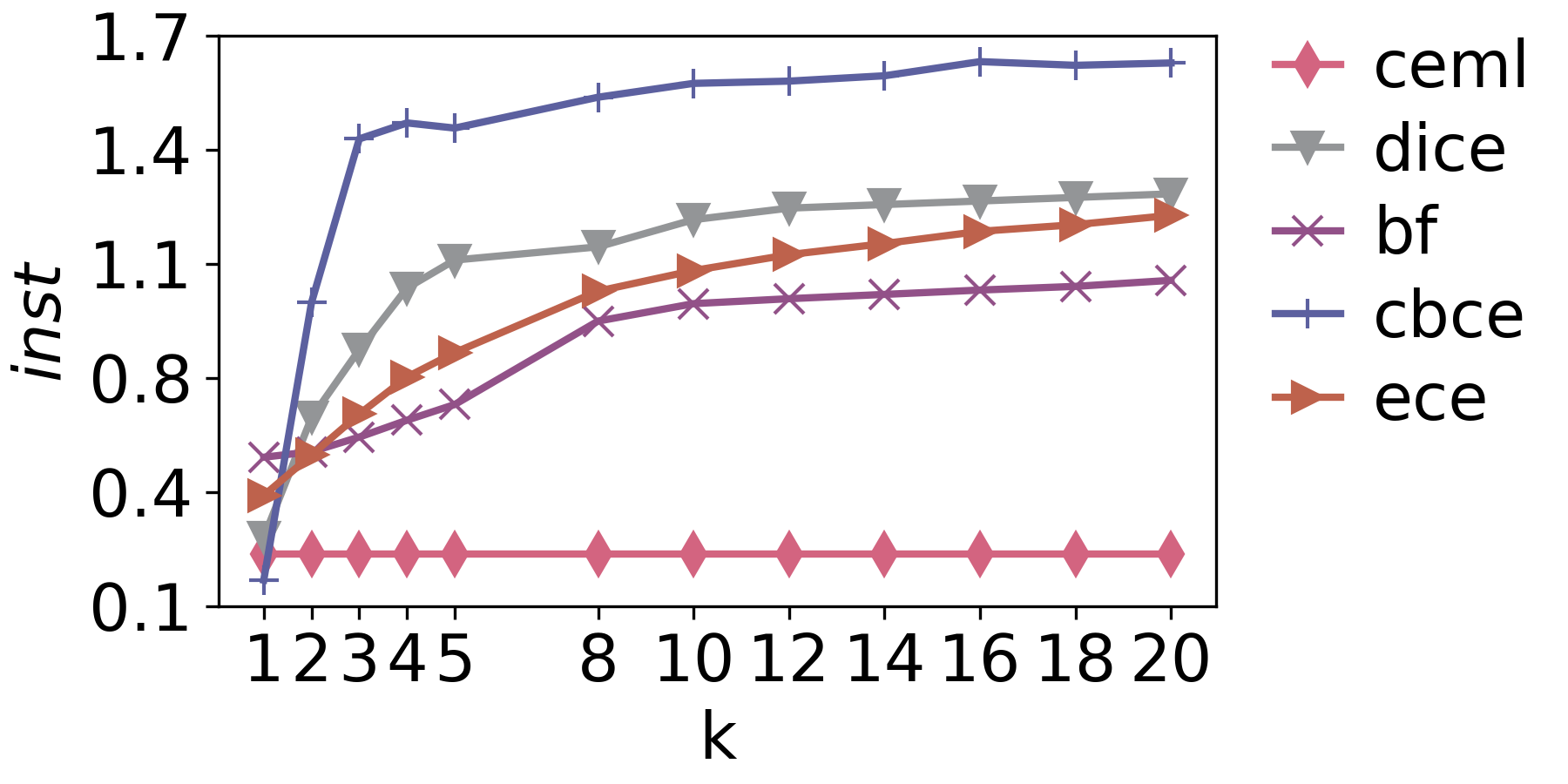}
    \hspace{-0.65cm}\includegraphics[trim = 0mm 0mm 0mm 0mm, clip,width=0.24\linewidth]{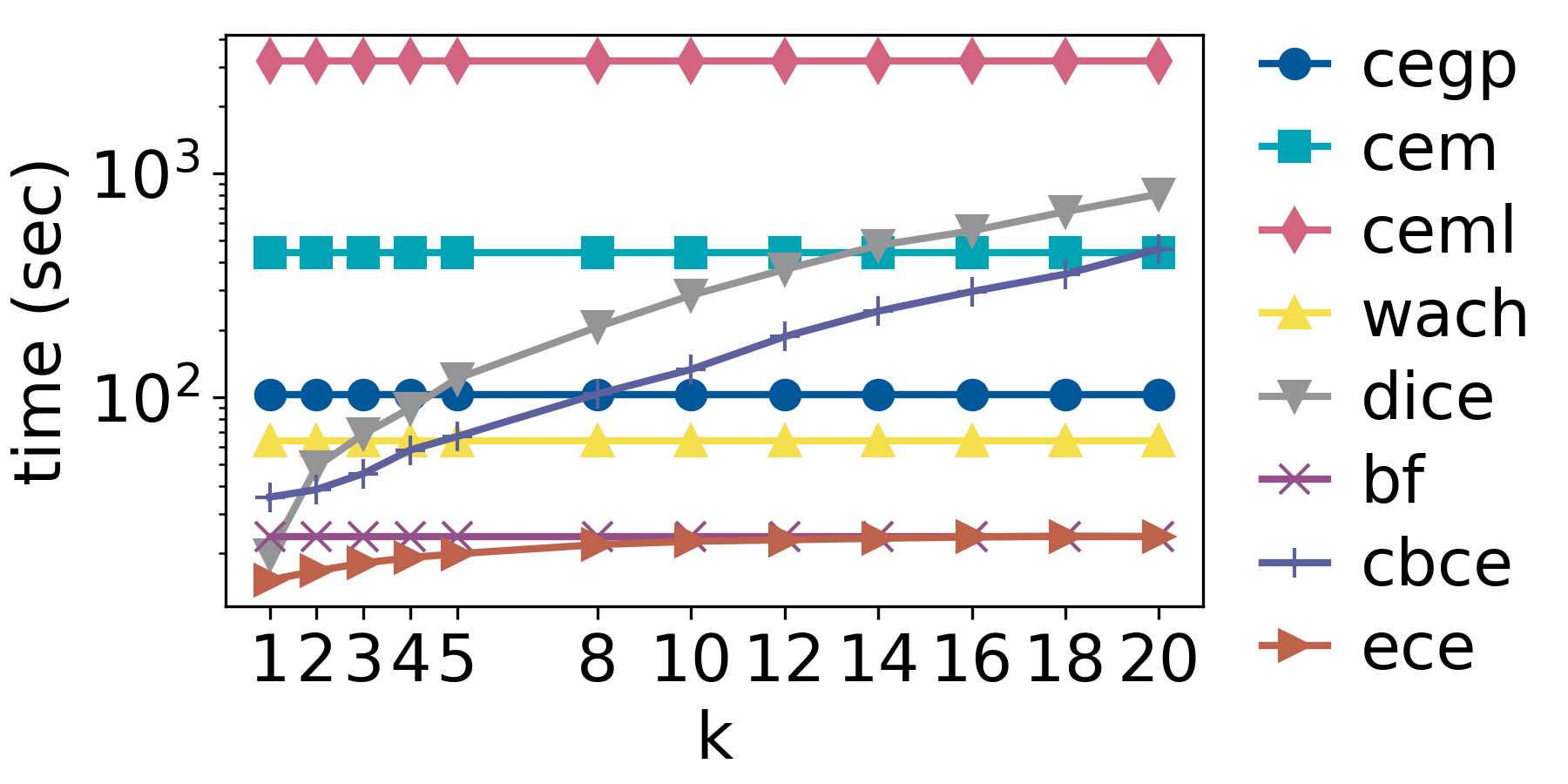}
    \caption{Aggregate metrics on tabular datasets by varying $k$.}
    \label{fig:competitors_var_k}
\end{figure*}

\begin{figure*}[t]
    %\centering
    \noindent \hspace{-0.55cm}
    \includegraphics[trim = 0mm 0mm 0mm 0mm, clip,width=0.25\linewidth]{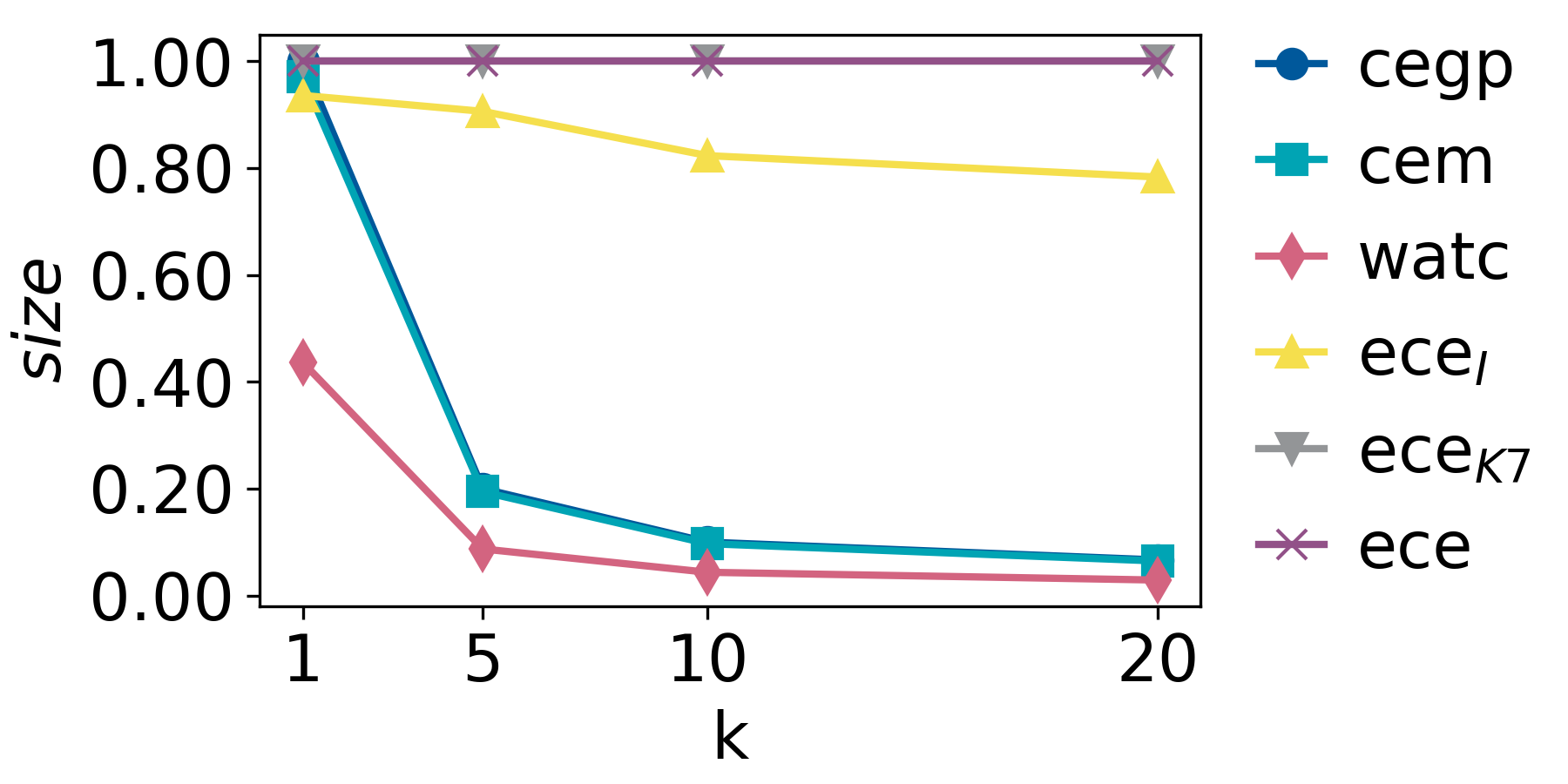}%
    \hspace{-0.65cm}\includegraphics[trim = 0mm 0mm 0mm 0mm, clip,width=0.25\linewidth]{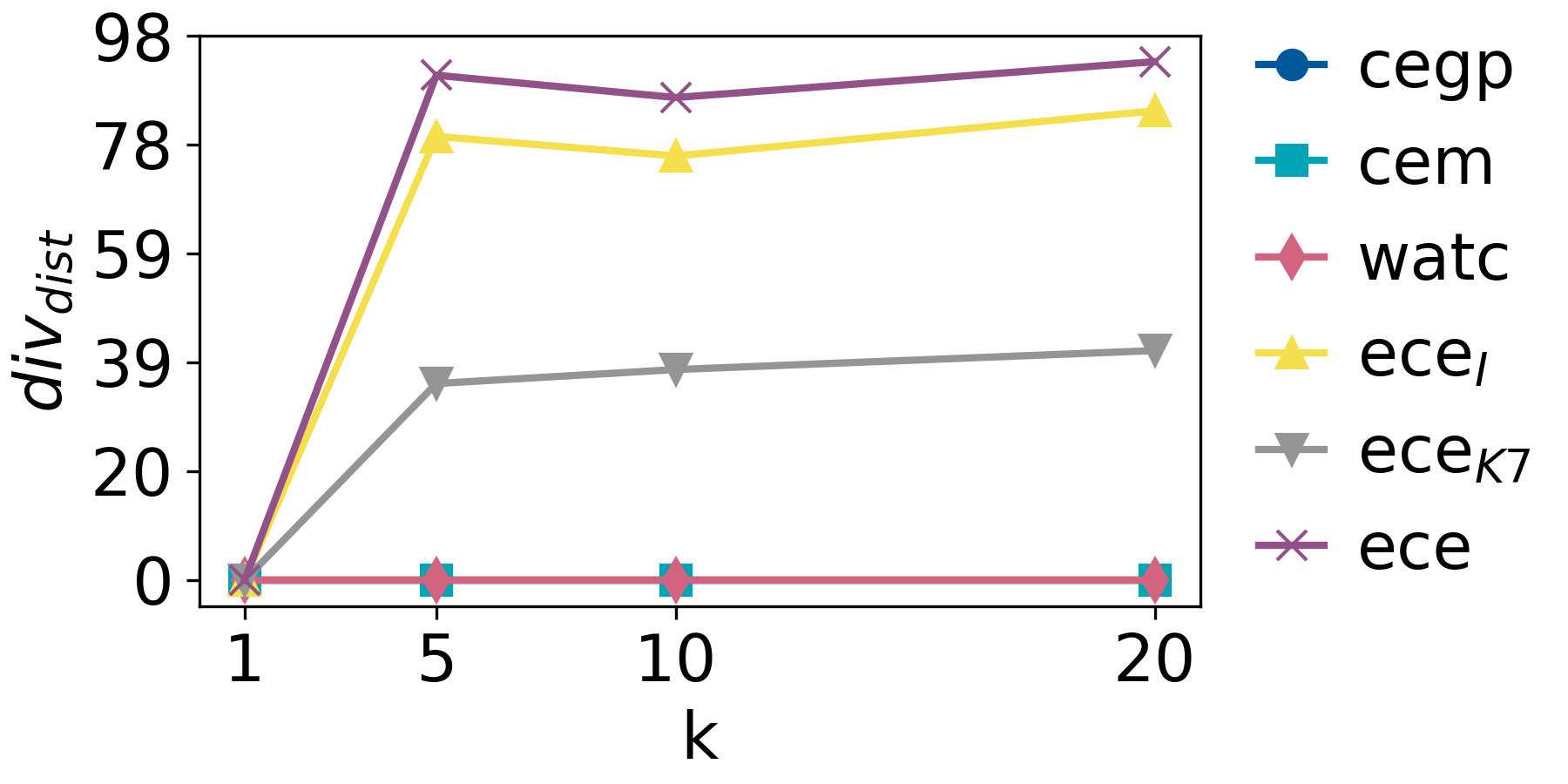}%
    \hspace{-0.65cm}\includegraphics[trim = 0mm 0mm 0mm 0mm, clip,width=0.25\linewidth]{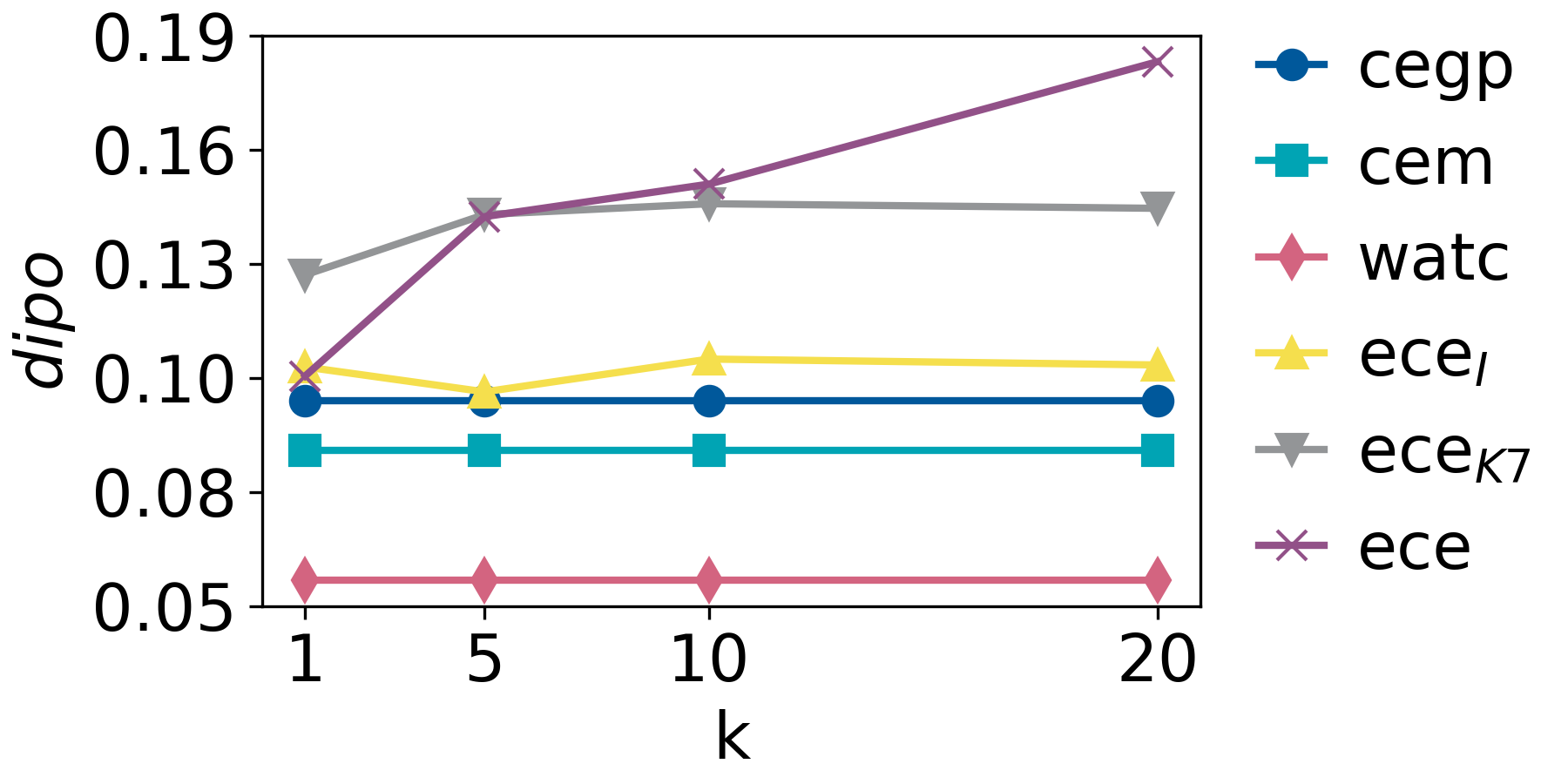}%
    \hspace{-0.65cm}\includegraphics[trim = 0mm 0mm 0mm 0mm, clip,width=0.25\linewidth]{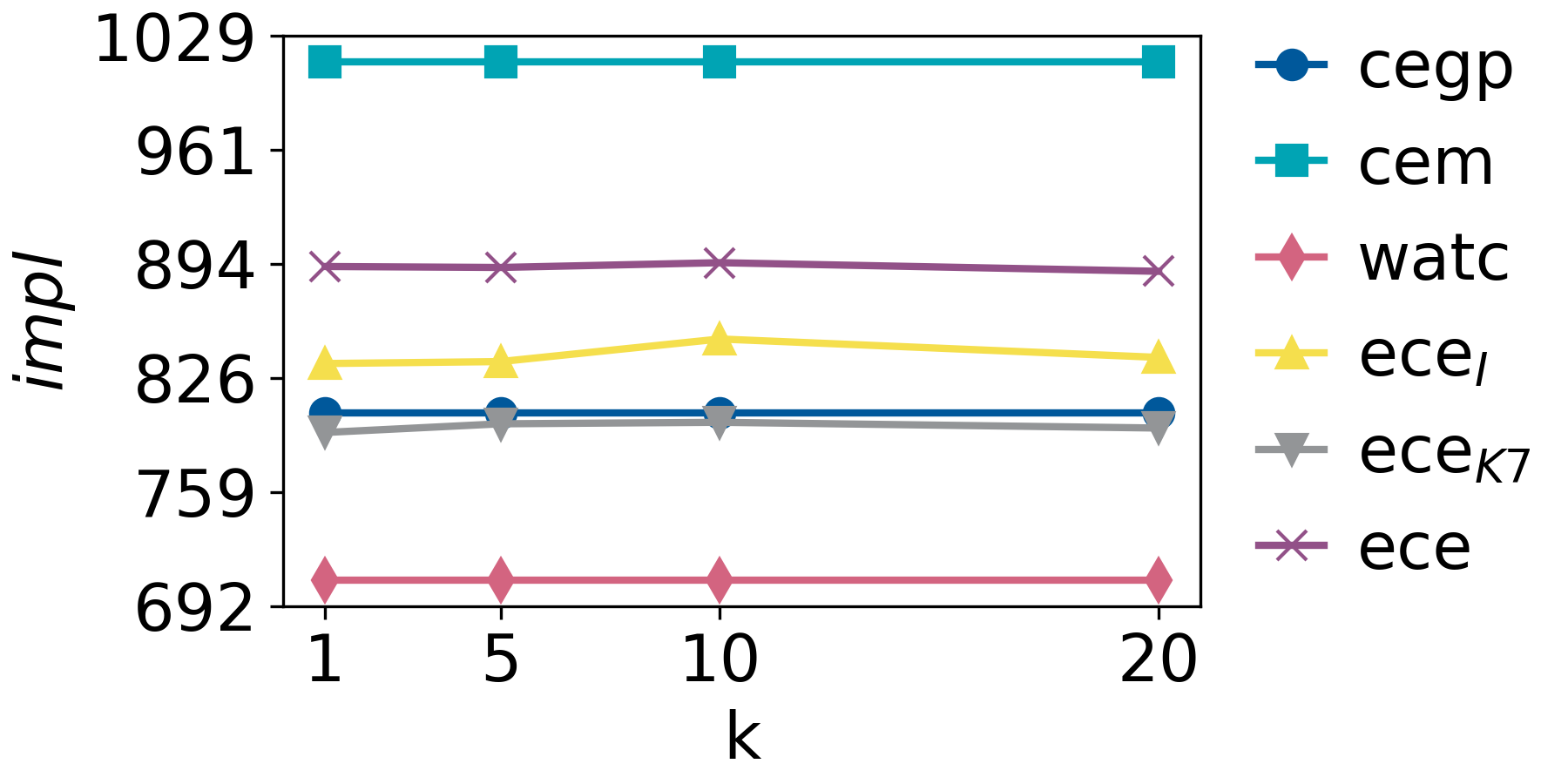}%
    \hspace{-0.65cm}\includegraphics[trim = 0mm 0mm 0mm 0mm, clip,width=0.25\linewidth]{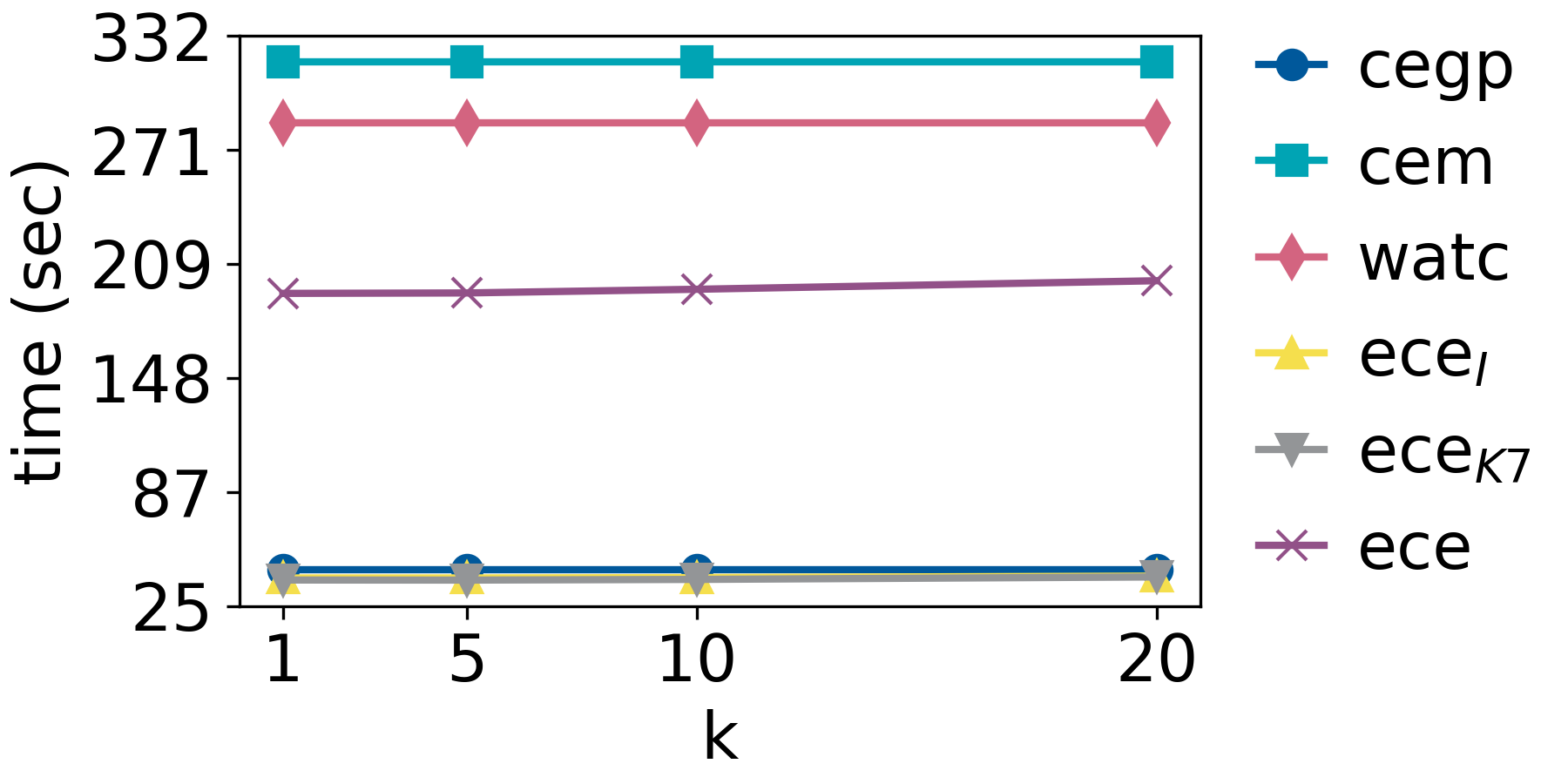}
    
    \hspace{-0.55cm}
    \includegraphics[trim = 0mm 0mm 0mm 0mm, clip,width=0.25\linewidth]{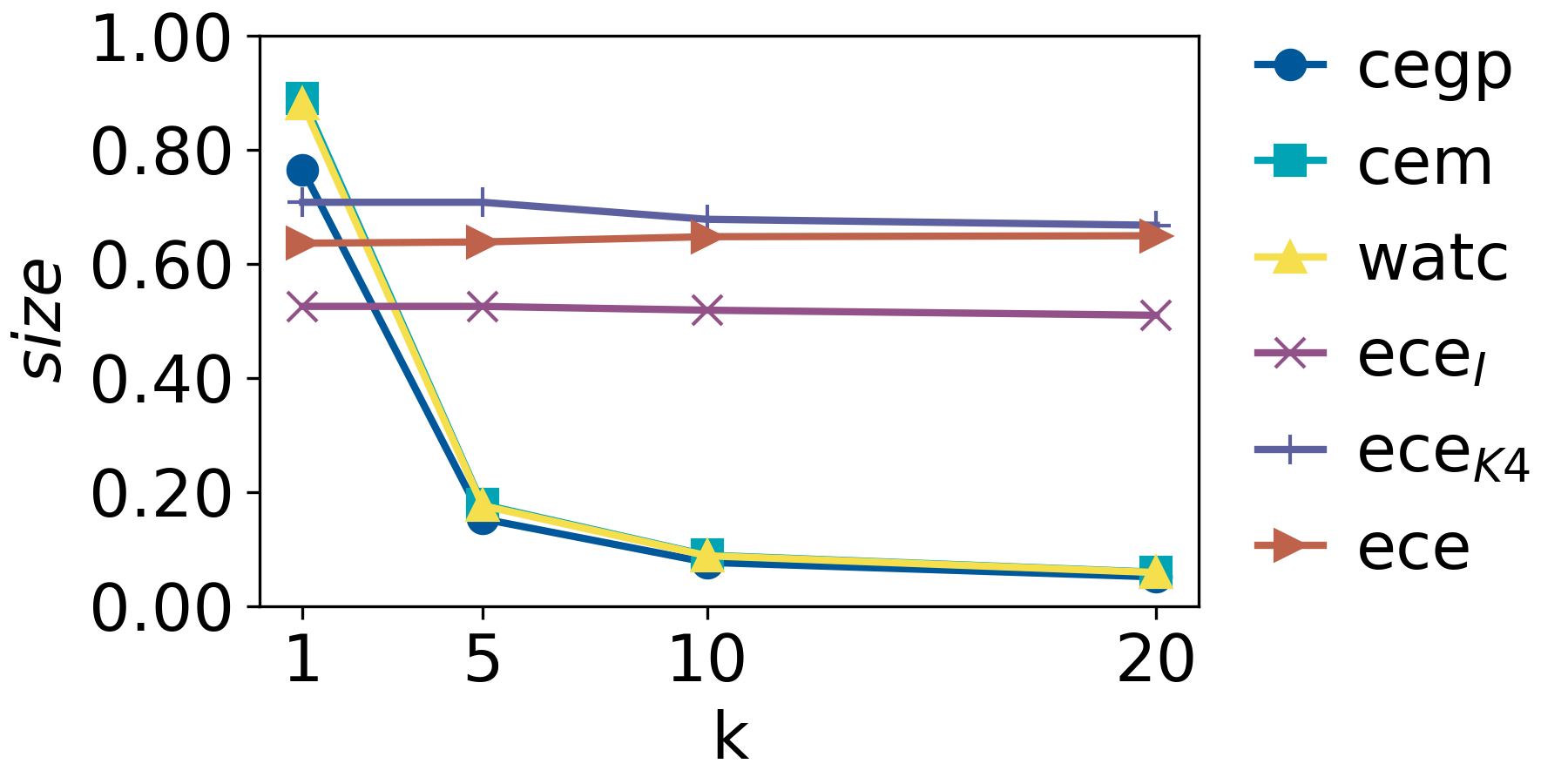}%
    \hspace{-0.65cm}\includegraphics[trim = 0mm 0mm 0mm 0mm, clip,width=0.25\linewidth]{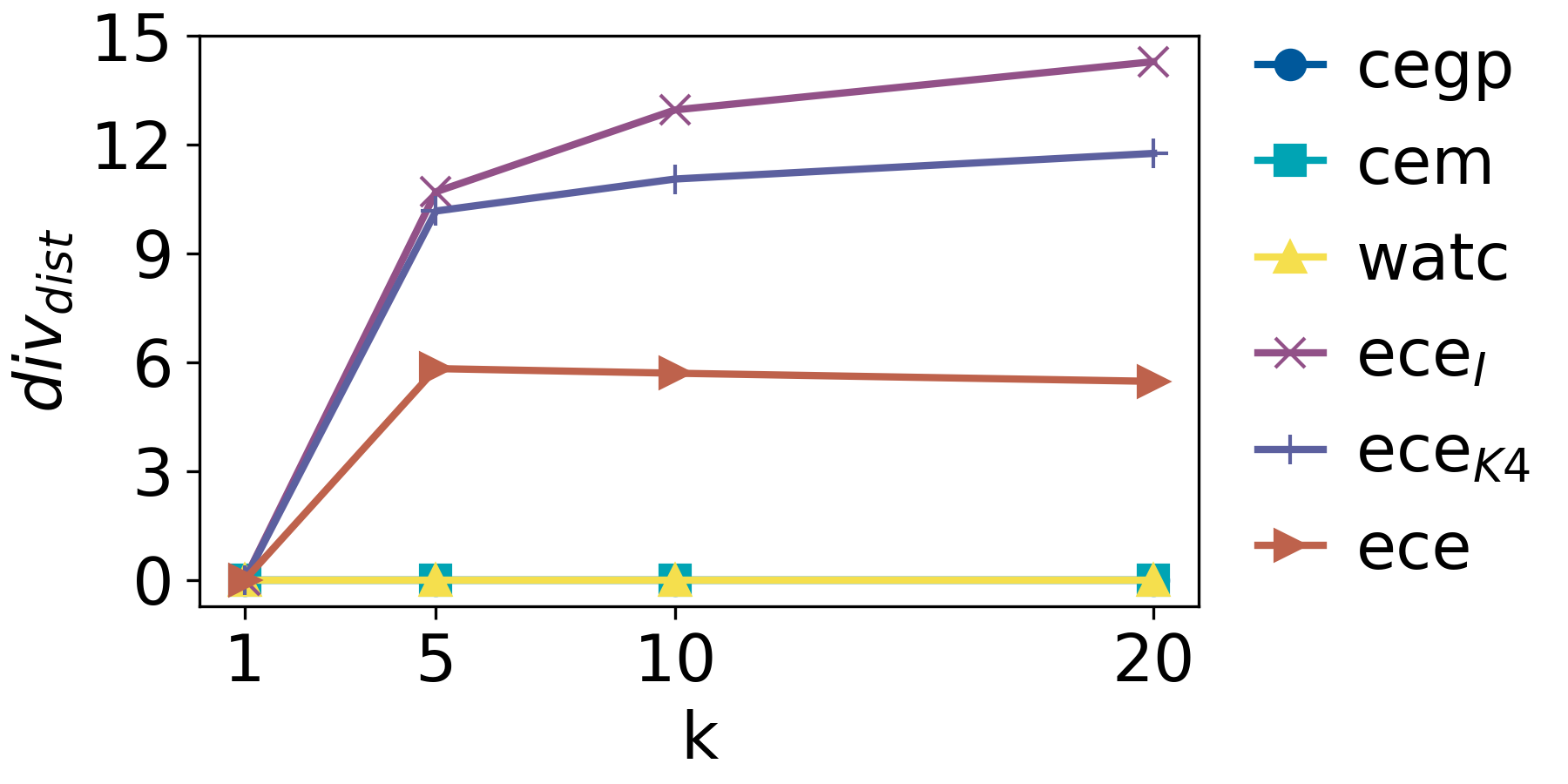}%
    \hspace{-0.65cm}\includegraphics[trim = 0mm 0mm 0mm 0mm, clip,width=0.25\linewidth]{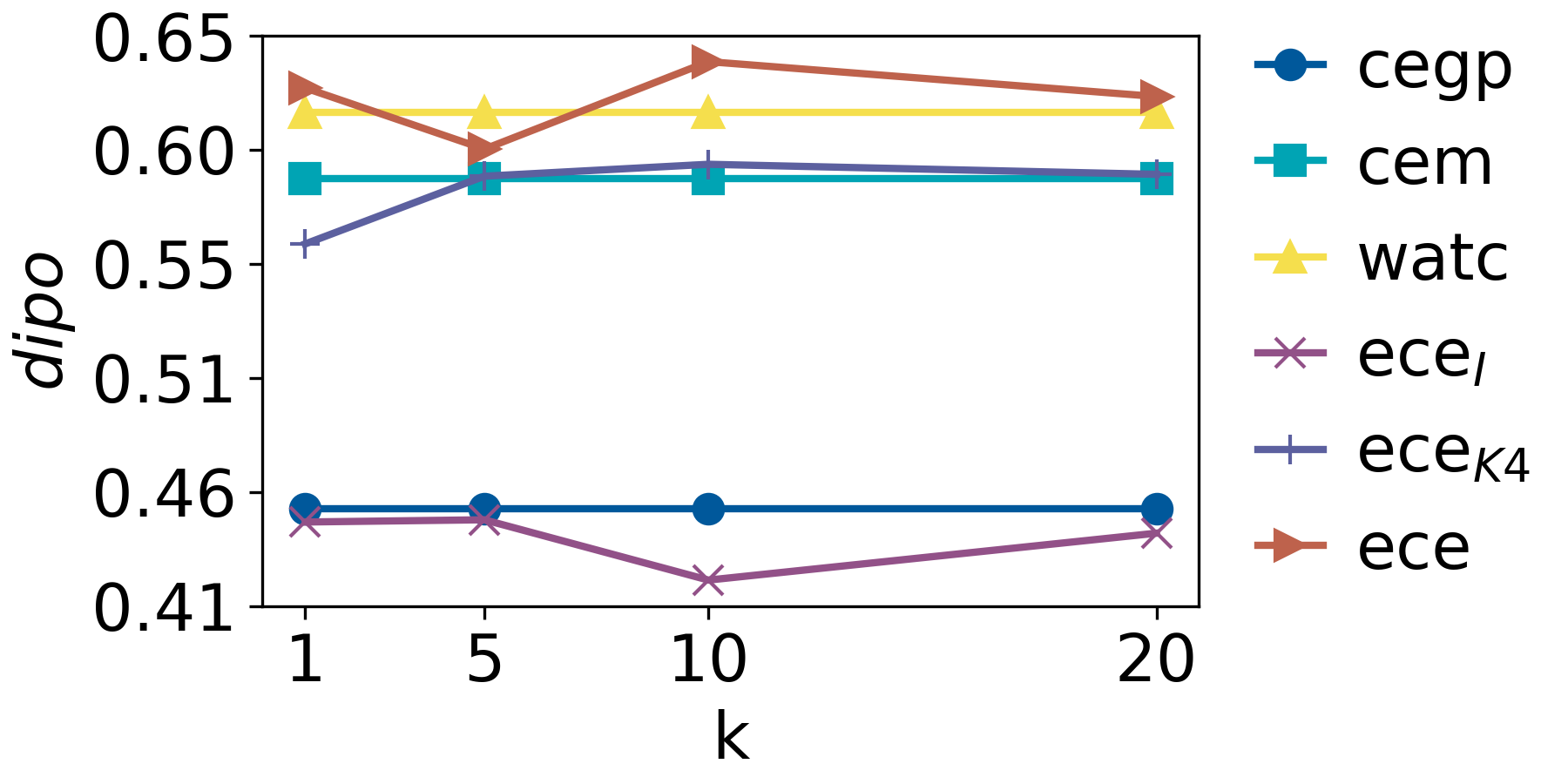}%
    \hspace{-0.65cm}\includegraphics[trim = 0mm 0mm 0mm 0mm, clip,width=0.25\linewidth]{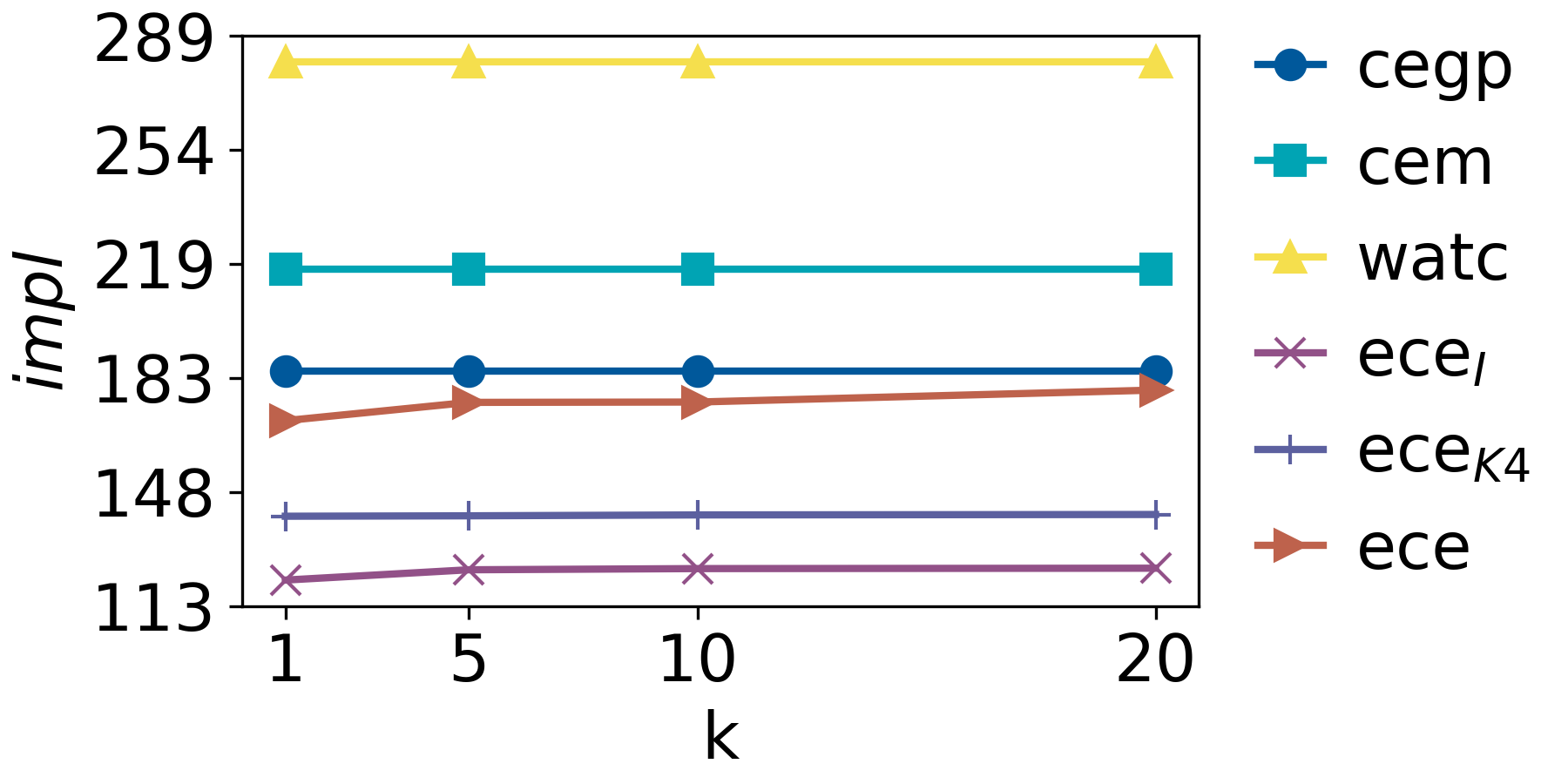}%
    \hspace{-0.65cm}\includegraphics[trim = 0mm 0mm 0mm 0mm, clip,width=0.25\linewidth]{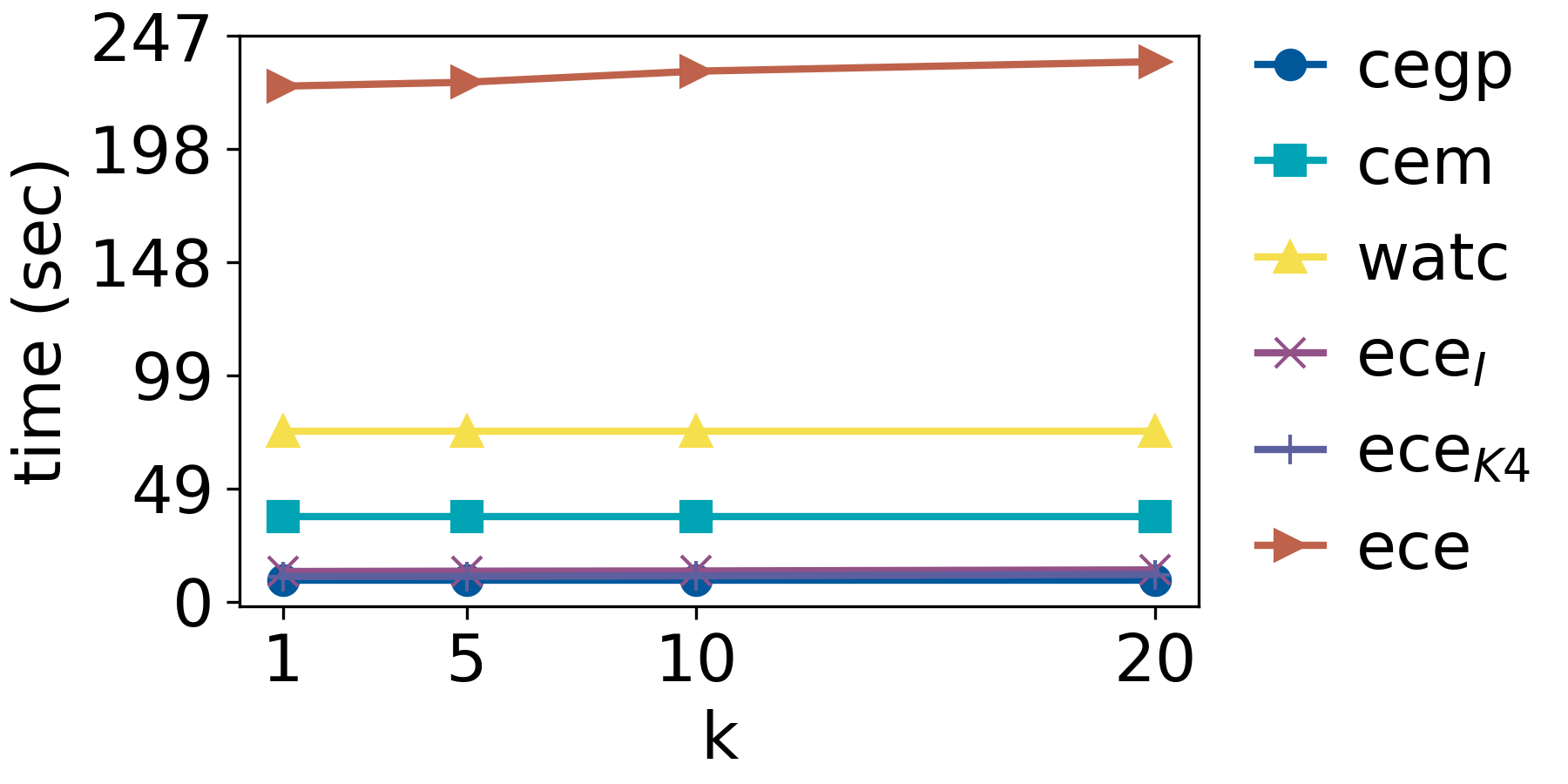}
    \caption{Aggregate metrics on images ($1^{\mathit{st}}$ row) and time series ($2^{\mathit{nd}}$ row) by varying $k$.}
    \label{fig:competitors_other_data_types}
\end{figure*}

\begin{figure}[t]
    \centering
    \includegraphics[trim = 20mm 0mm 20mm 0mm, clip,width=0.33\linewidth,valign=t]{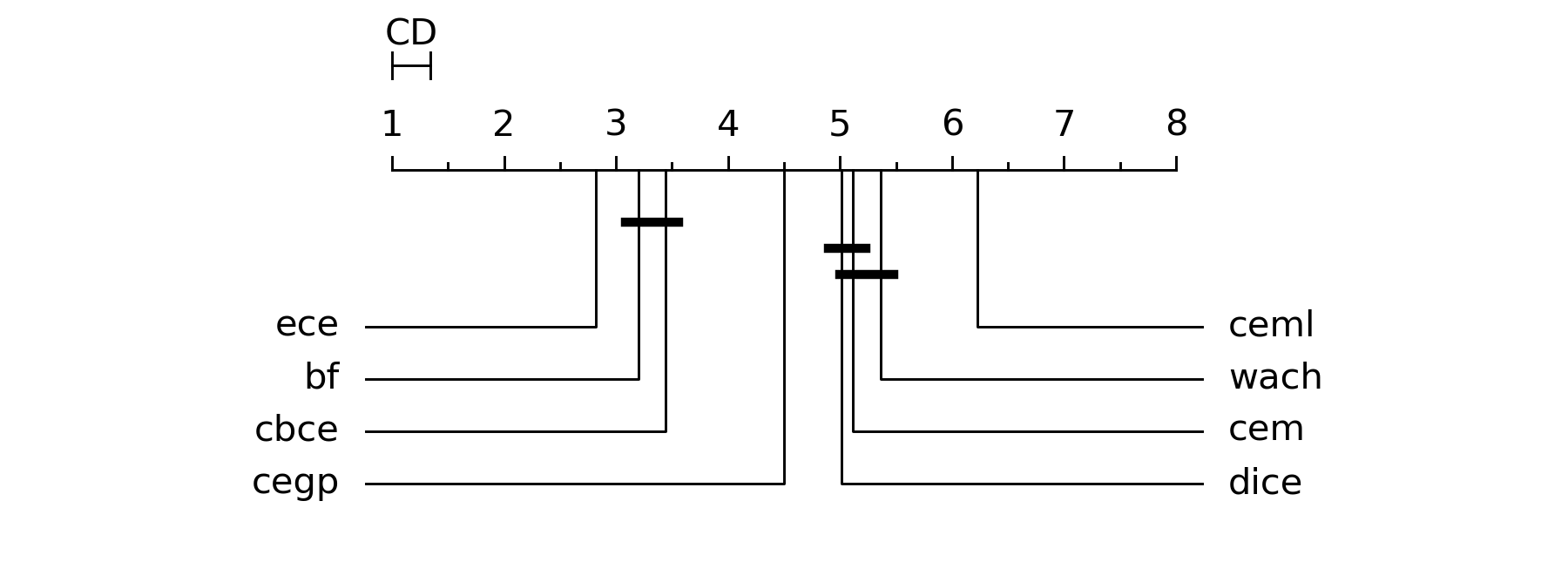}%
    \includegraphics[trim = 20mm 0mm 20mm 0mm, clip,width=0.33\linewidth,valign=t]{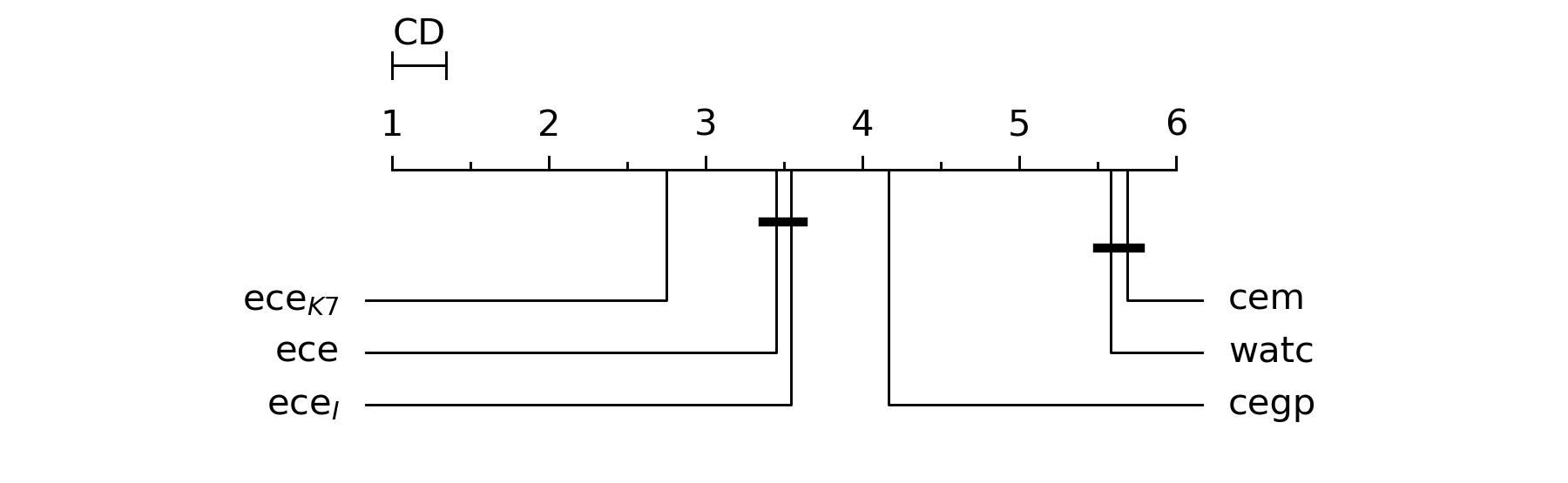}%
    \includegraphics[trim = 20mm 0mm 20mm 0mm, clip,width=0.33\linewidth,valign=t]{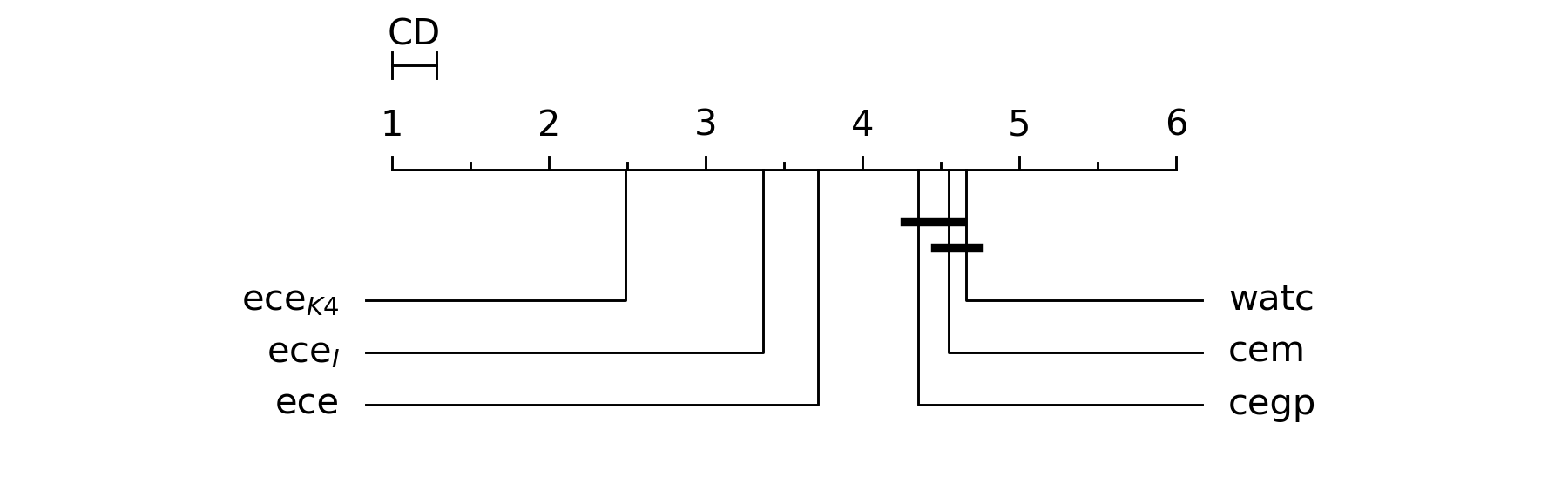}
    \caption{Critical Difference (CD) diagrams for the post-hoc Nemenyi test at 95\% confidence level: tabular (left), images (center), and time series (right) datasets.}
    \label{fig:cd_plot}
\end{figure}

\medskip {\bf Quantitative Evaluation.}
Fig.~\ref{fig:competitors_var_k} shows the performance of the compared explainers on tabular data when varying $k$.
From the first plot, we notice that only \rce{}, \dice{}, \cbce{} and \brfo{} are able to return at least 80\% of the required counterfactuals. Most of the other methods only return a single one.
From the second plot, we conclude that only \rce{}, \brfo{} and \cbce{} return a notable fraction of actionable counterfactuals ($\mathit{act}$).
%\dice{} and \ceml{} return more than one counterfactual and they allow to specify the actionable features, but they do not actually check their results for validity or actionability. 
From the plots on dissimilarity ($\mathit{dis}_{count}$ and $\mathit{dis}_{dist}$) and diversity ($\mathit{div}_{count}$ and $\mathit{div}_{dist}$), it turns out that \cbce{} (and also \dice{}) has good values of diversity, but performs poorly w.r.t.~dissimilarity. \brfo{} wins over \rce{} w.r.t.~the $\mathit{dis}_{dist}$ measure, loses w.r.t.~the $\mathit{div}_{dist}$ measure, and is substantially equivalent w.r.t.~the other two measures.
As for discriminative power $\mathit{dipo}$, \rce{} performs slightly lower than \dice{}, \cbce{}, \brfo{} and \ceml{}.
Regarding plausibility ($\mathit{impl}$), \rce{} is the best performer if we exclude methods that return a single counterfactual (i.e.,~\cem{}, \cegp{} and \watc{}).
Indeed, \rce{} $\mathit{impl}$ is constantly smaller that \dice{} and \brfo{} and in line with \cbce{}, which is the only endogenous methods compared.
Intuitively, counterfactuals returned by \rce{} resemble instances from the reference population.
Concerning instability $\mathit{inst}$, \rce{} is slightly worse than \brfo{} and slightly better than \dice{}.
\ceml{} is the most stable, and \cbce{} the most unstable.
\cem{}, \cegp{} and \watc{} are not shown in the instability plot because, in many cases, they do not return counterfactuals for both of the two similar instances. 
Finally, all the explainers, with the exception of \brfo{} and \rce{}, require on average a runtime of more than one minute. 
%\textcolor{green}{We do not report the results of experiments observing the effect of varying the size of the training set $|X|$ on the various evaluation measures either because of lack of space but also because none of the algorithm seems to be particularly affected by working on a smaller sample of data up to a minimum of the 20\% of $|X|$.}
We summarize the performances of the approaches by the CD diagram in Fig.~\ref{fig:cd_plot}~(left), which shows the mean rank position of each method over all experimental runs (datasets $\times$ black boxes $\times$ metrics $\times$ $k$). Overall, \rce{} performs better than all competitors, and the difference is statistically significant. %: the simple \brfo{} explainer has comparable performances, but with lower diversity.

Fig.~\ref{fig:competitors_other_data_types} shows the performance on images (first row) and time series (second row) datasets.
We consider also the \rce{} with the identity encoder/decoder (named \rce{}$_I$), and with the kernel encoder/decoder (\rce{}$_ {K7}$ for kernel of size $7 \times 7$ and \rce{}$_ {K4}$ for kernel of size $4 \times 4$).
For images, \cem{}, \cegp{} and \watc{} return only a single counterfactual, while \rce{} provides more alternatives and with the best diversity.
\watc{} returns the least implausible counterfactuals, the variants of \rce{} stand in the middle, while \cem{} returns less realistic counterfactuals.
Regarding running time, \cegp{} is the most efficient together with \rce{}$_I$ and \rce{}$_{K4}$.
The usage of the autoencoder in \rce{} increases the runtime.
\cem{} and \watc{} are the slowest approaches.
Similar results are observed for time series, with few differences.
%All the approaches find less counterfactuals. 
%\rce{} returns less diversified sets than \rce{}$_I$ and \rce{}$_K$ (see $\mathit{div}$ plot).
%\rce{}$_I$ and \rce{}$_K$ return the most plausible sets of counterfactuals and are the two most efficient approaches. 
The CD diagrams in Fig.~\ref{fig:cd_plot}~(center, right) confirm that \rce{} and its variants are the best performing methods.

%\section{Conclusions}
%\label{sec:conclusion}
%We have proposed a model/data agnostic ensemble framework for counterfactual explanation of black box decisions.
%\rce{} selects counterfactuals from the results of base $k$-counterfactual explainers invoked over sample 
%instances and features.  Simple base explainers return valid and actionable counterfactuals. 
%Selection of counterfactual favors diversity. 
%Experimental evaluation has shown that \rce{} achieves good to best performances w.r.t.~several metrics.

\medskip
{\bf Acknowledgment.}
Work partially supported by the European Community H2020-EU.2.1.1 programme under the G.A. 952215 \emph{Tailor}. 

\bibliographystyle{splncs04}
%\bibliography{biblio_short}

\end{document}